\documentclass{article}

\usepackage[preprint]{neurips_2026}

\usepackage[utf8]{inputenc}
\usepackage[T1]{fontenc}
\usepackage{amsfonts}
\usepackage{amsmath}
\usepackage{amssymb}
\usepackage{hyperref}
\usepackage{url}
\usepackage{booktabs}
\usepackage{microtype}
\usepackage{xcolor}
\usepackage{graphicx}
\usepackage{enumitem}

\hypersetup{
  colorlinks=true,
  linkcolor=blue!50!black,
  citecolor=blue!50!black,
  urlcolor=blue!50!black,
}

\graphicspath{{figures/}}

\title{Protein contacts are already in the attention: \\
a single-forward-pass alternative to the Categorical Jacobian}

\author{%
  Rome Thorstenson\\
  Independent Researcher\\
  \texttt{rome.thorstenson@aya.yale.edu}\\
}

\begin{document}
\maketitle

\begin{abstract}
\noindent
The Categorical Jacobian of \citet{zhang2024cj} reads protein contacts
from a language model by perturbing every residue with every alternative
amino acid, about $19L$ forward passes. We show the signal it reconstructs
is already concentrated in a small subset of attention heads: averaging
the top-$K$ contact-relevant heads --- selected on as few as 10 labeled
proteins, with no fitted per-pair or per-head weights --- recovers
contacts in a single forward pass and matches or beats the Categorical
Jacobian for every bidirectional model where it is defined (bar the
smallest, 8M). Our primary test is
leakage-clean: on a CAMEO split where neither selection nor evaluation
touches data the models have plausibly memorized, the head readout beats
the Categorical Jacobian on ESM-2-650M by $+9$pp ($N{=}29$, $p<0.001$),
with the within-model margin reproducing across architectures. Ablations
localize the gain to labeled head \emph{selection}, not to averaging: at
a matched label budget the unweighted mean ties a supervised $L_1$
logistic regression on the same heads. Both methods fall 30--36pp from
their in-distribution Zhang numbers to the leakage-clean split, which we
read as an upper bound on how much prior numbers reflect pretraining
overlap. We additionally introduce representation-CJ, a hidden-state
generalization of the Jacobian to architectures without a masked-LM head
(the output-head-independent analogue of logit-CJ), agreeing with the
Categorical Jacobian where both are defined (per-protein Pearson $r \approx 0.95$); show that
the optimal $K$ tracks how diffusely a model spreads its contact heads;
and find both methods lose the signal on the two causal LMs we test,
suggesting attention-encoded pair structure may depend on bidirectional
pretraining.\footnote{Code, figure-regeneration
scripts, and the summary result tables (per-cell means and bootstrap
intervals) are available at
\url{https://github.com/Rome-1/plm-contact-fusion}.}
\end{abstract}

\section{Introduction}

Protein language models (PLMs) trained on hundreds of millions of
unlabeled sequences learn rich representations of protein structure
\citep{rives2021esm1b,lin2023esm2}. A natural question follows: can
the structural knowledge a PLM acquires during pretraining be read out
\emph{without any supervised training on structure labels}? One
answer, proposed by \citet{zhang2024cj}, is the Standard Categorical
Jacobian (CJ): perturb each residue position with every alternative
amino acid, measure the response in the model's per-position masked-LM
logits, and read residue--residue contacts off the resulting Jacobian.
CJ reaches $\text{P}@L \approx 0.80$ on a 1{,}431-PDB-chain benchmark
using ESM-2-650M and requires no contact labels at all. However, the
construction takes $\sim 19L$ forward passes for a protein of $L$
residues (one perturbation per residue $\times$ 19 alternative amino
acids) and is structurally tied to a per-position categorical head,
which restricts it to bidirectional masked-LM architectures.

This paper offers a simpler answer: the same contact signal already
sits in the attention maps of the model, concentrated in a
small, identifiable subset of heads. Intuitively, residues that are
close in three-dimensional space exert strong evolutionary constraints
on each other --- substituting one often requires a compensating change
at the other to preserve the fold. Masked-LM pretraining captures
these pairwise dependencies implicitly, and the attention mechanism is
the natural site for the model to represent which positions ``attend
to'' which. \citet{bhattacharya2021attention} showed on ProtBERT-BFD
that 6 of 480 attention heads carry most of the contact-relevant
signal; \citet{rao2021transformerplm} reached a related conclusion on
ESM-1b via an $L_1$-sparse supervised regression that effectively
selects a small subset of heads, and found that simply averaging those
heads unweighted outperforms the fitted regression. What is new here
is not the mean --- \citet{rao2021transformerplm} already showed it
beats their fitted regression --- but that it needs no regression and
no per-pair labels (a 10-protein ranking suffices), survives a
leakage-clean evaluation, improves on the Categorical Jacobian, and
extends to encoder--decoder architectures (it can be applied to causal
LMs too, but the signal collapses there; Section~\ref{sec:causal}).
We extend this line in a label-efficient direction. The unweighted arithmetic mean of
the top-$K$ heads, with $K$ chosen once from a disjoint 10-protein
head-precision ranking and no per-pair weight fitting, gives a
label-efficient contact-prediction signal that exceeds CJ on every masked-LM
and encoder--decoder PLM we test, in one forward pass instead of
$\sim 19L$, on data filtered to remove sequence-level pretraining
overlap. The
optimal $K$ varies between architectures and tracks the diffuseness of
the contact-head cluster: a mechanistic observation about how different
PLMs allocate pair structure across their attention heads.

\begin{figure}[t]
  \centering
  \includegraphics[width=0.95\linewidth]{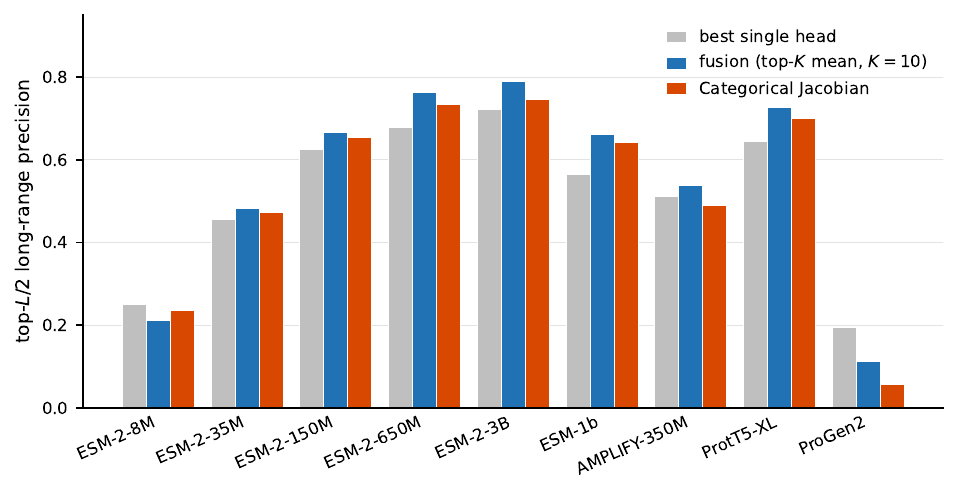}
  \caption{Top-$L/2$ long-range precision (fraction of the $L/2$
  highest-scoring $|i{-}j|\geq24$ pairs that are true contacts) on Zhang
  eval-200 across nine
  protein-LM variants (ESM-2 at five scales, ESM-1b, AMPLIFY-350M,
  ProtT5-XL, ProGen2-xlarge). Naive-mean fusion of the top-$K$ attention
  heads (blue) beats the per-protein top-1 head (grey) and the
  best-available Categorical Jacobian (red) on every cell at and above
  the 35M scale; the concentrated-cluster ESM-2-8M (best read at
  $K{=}3$, not the $K{=}10$ plotted here) is the in-distribution
  exception, and ProGen2-xlarge is the causal-LM scope boundary.
  ``Best-available CJ'' is logit-CJ \citep{zhang2024cj} where
  defined (ESM-2 family, ESM-1b) and repr-CJ
  (Section~\ref{sec:repr-cj-method}) where logit-CJ does not apply
  (AMPLIFY, ProtT5, ProGen2); the scope boundary is detailed in
  Section~\ref{sec:causal}. Per-protein 95\% bootstrap
  intervals in Appendix Table~\ref{tab:a1}.}
  \label{fig:headline}
\end{figure}

\paragraph{Contributions.}
The end-to-end leakage-clean comparison is the central finding
(\emph{leakage-clean}: evaluated only on protein chains deposited after a
tested model's pretraining cutoff and filtered for sequence similarity
per \citet{hermann2024leakage}):
with head selection on one half of CAMEO-PTA25 (our 2025 pretraining-aware
CAMEO split of post-2022-09 targets, defined in Section~\ref{sec:setup})
and evaluation on the
disjoint other half (no Zhang chains anywhere in the pipeline),
fusion on ESM-2-650M beats CJ, and CJ no longer beats the best single
attention head (Section~\ref{sec:fusion-beats-cj}). The pattern
reproduces on a wider CAMEO-PTA25 set across three
architectures (ESM-2-650M, ESM-2-3B, ProtT5-XL). Three further results
follow. \emph{(i)~Cross-architecture replication} on Zhang eval-200 and
CASP14 free-modeling (CASP14-FM) extends to five PLMs spanning masked-LM and encoder--decoder
objectives. \emph{(ii)~Representation-CJ}, a hidden-state
generalization of logit-CJ, applies to architectures lacking a
per-position MLM head and agrees with logit-CJ where both are defined
(Section~\ref{sec:repr-cj}). \emph{(iii)~Causal-LM scope}: both fusion
and CJ collapse on both causal LMs we tested (ProGen2-xlarge and
ProGen2-large), consistent with bidirectional pretraining being important
for attention-encoded pair structure --- though with five architectures
and several correlated confounds we report this as associational, not
causally identified (Section~\ref{sec:causal}). Both
methods drop 30--36pp absolute from Zhang eval-200 to leakage-clean CAMEO,
upper-bounding how much of the prior unsupervised-contact-prediction
headline reflects sequence overlap between pretraining and evaluation
(a CAMEO difficulty shift contributes too).

\section{Related Work}
\label{sec:related}

\paragraph{Unsupervised contact prediction from protein language models.}
\citet{zhang2024cj} introduced the Standard Categorical Jacobian:
perturb each residue with every alternative amino acid, measure the
per-position categorical-distribution response, and read contacts off
the resulting Jacobian. Their primary result is
$\text{P}@L = 0.80$ on a 1{,}431-PDB-chain benchmark with ESM-2-650M,
above an inverse-covariance baseline of 0.61.
We adopt CJ as the direct unsupervised baseline this paper compares
against and we reproduce it at
$\text{P}@L/2 \text{ long} = 0.733$ on our 200-protein eval draw
(head/$K$ selected on a disjoint 10-protein set) of
their benchmark (the metric switch is for direct comparability with
the attention-readout literature
\citep{rao2021transformerplm,rao2021msa}, which reports
top-$L/2$ long-range precision). Our fusion method extends this
unsupervised-contact-prediction line --- read contact signal off a
pre-trained PLM without a supervised head --- but reads attention
rather than perturbing the input, exchanging $\sim 19L$ forwards
for 1 forward at higher precision. Representation-CJ (next
paragraph) is a parallel methodological piece.

\paragraph{Attention as a contact signal.}
The structural localization of pair information in transformer
attention was first observed in NLP \citep{vig2019bertattention,clark2019bertlookat}
and extended to PLMs by \citet{vig2021bertology}, \citet{bepler2019contact},
\citet{rao2021transformerplm}, \citet{rao2021msa}, and
\citet{bhattacharya2021attention}. The closest methodological neighbor
is \citeauthor{rao2021transformerplm}'s $L_1$-regularized logistic
regression over per-head average-product-corrected (APC'd) attention maps on ESM-1b, which acts
as an implicit sparse head selector ($\sim 102/660$ heads receive
non-zero weight; top-$L/2$ long-range $= 0.533$); they additionally
report that averaging the top-10 selected heads \emph{unweighted}
outperforms the fitted regression. Our setup is closest in mechanism
to that unweighted top-$k$ variant: we differ in selecting on a
small head-precision ranking (as few as 10 proteins) rather than via a
regression fit to per-pair contact labels on full PDB chains. \citet{rao2021msa}
extends the recipe to multiple-sequence-alignment (MSA) inputs at higher precision.
\citet{bhattacharya2021attention} demonstrated, on \emph{ProtBERT-BFD},
that contact signal concentrates in a small subset of heads (6 of
480) and that a single layer's attention is often sufficient; the
head-concentration phenomenon transfers to the ESM family in our
cross-architecture sweep (Section~\ref{sec:ksweep}).
This prior work raises two questions we take up in the results: whether
the gain comes from head \emph{selection} or from averaging \emph{per
se}, and whether a fitted regression beats an unweighted mean at a
matched label budget. Both are answered in
Section~\ref{sec:fusion-beats-cj} and Appendix~\ref{app:rao-baseline}
(in short: selection is the operation that matters, and at a 50-protein
budget the fitted regression and our naive mean are statistically
indistinguishable).

\paragraph{Mechanistic interpretability of PLMs.}
A recent line of work uses sparse autoencoders and circuit analysis
to interpret what protein language models compute
\citep{adams2025plmsae,simon2025interplm}, identifying features for
binding sites, structural motifs, and functional domains in ESM-2
embeddings. These papers preserve CJ as their accuracy metric and
ask which model latents drive it. Our work is methodological: we
move accuracy and compute. The ``head cluster'' structure we exploit
in Section~\ref{sec:ksweep} is consistent with mech-interp findings
that contact information localizes to a small set of attention heads.

\paragraph{Leakage as a confound in PLM evaluation.}
\citet{hermann2024leakage} showed that splits drawn before the
pretraining cutoff inflate measured PLM downstream-task performance
by 11.1\% absolute on thermostability prediction; a pretraining-aware
split (filtering test sequences against pretraining cluster
representatives) recovers the model's true generalization.
\citet{bartoszewicz2025ppileakage} extend the same finding to
protein--protein interaction inference. Earlier PLM benchmark suites
TAPE \citep{rao2019tape} and FLIP \citep{dallago2021flip} adopted
identity-based holdout splits in the same spirit, and
\citet{detlefsen2022plmrepr} flagged pretraining-data overlap as a
confound for hidden-state-based PLM readouts more broadly.
\citet{liveproteinbench2025} introduces a contamination-free
benchmark whose test set is restricted to proteins released after 2024. The structural-prediction literature has
its own temporal-holdout protocols: CASP rounds release targets after
the relevant pretraining cutoffs \citep{kryshtafovych2021casp14}, and
AlphaFold2's training/test split is chronologically separated by
design \citep{jumper2021af2}. What is novel in our application is the
specific Hermann-filter evaluation of the CJ-vs-attention-readout
comparison for unsupervised contact prediction, with parameters chosen
to match \citeauthor{hermann2024leakage}'s sequence-similarity
protocol (Section~\ref{sec:fusion-beats-cj}).

\paragraph{Generalization across architectures.}
\citet{zhang2024cj} evaluate CJ exclusively on the ESM family. We
extend the evaluation to an encoder-decoder model (ProtT5-XL;
\citealp{elnaggar2021prottrans}), bidirectional MLM variants beyond
ESM-2 (ESM-1b, \citealp{rives2021esm1b}; AMPLIFY-350M,
\citealp{fournier2024amplify}), and a causal LM (ProGen2-xlarge;
\citealp{madani2023progen2}); we introduce repr-CJ as the
technical change that makes the cross-architecture comparison
possible. We additionally evaluate ESM Cambrian (ESMC;
\citealp{esmteam2024esmc}), the bidirectional masked-LM successor to
ESM-2, distributed open-source as part of a public protein
``world model'' release \citep{biohub2026esm}: at both scales it
exposes a categorical logit head and per-head attentions, so logit-CJ
and fusion apply unchanged, and the within-model fusion-over-CJ margin
holds on it (Section~\ref{sec:fusion-beats-cj},
Table~\ref{tab:repr-cj-arch}). We leave the multi-track generative ESM-3
\citep{hayes2025esm3} --- whose generative architecture departs from
the masked-LM readout our methods assume --- to future work.

\paragraph{Supervised alternatives.}
Supervised structural prediction \citep{jumper2021af2,lin2023esm2}
reaches substantially higher absolute contact precision than any
attention-readout method, including ours. The comparison
is orthogonal: supervised methods learn a structure-prediction
function on labeled contact maps, while readout methods (CJ, and our
few-shot head selection) recover what the PLM has \emph{already} encoded
from sequence pretraining alone, using no supervised contact head. The question we ask here --- what does the model know
without further training? --- has a different epistemic status than
the question of how accurately a supervised head can be fit on top
of it. We comment in Section~\ref{sec:discussion} on the implicit
comparison.

\section{Methods}
\label{sec:methods}

\subsection{Notation and protocol}
\label{sec:setup}

A protein of $L$ residues folds into a three-dimensional structure in
which residues far apart in the chain can be close in space. A
\textbf{contact} is a pair of residues $(i, j)$ within $8$\,\AA{} in
the experimentally determined structure.\footnote{We follow the standard
criterion of each source benchmark: an $8$\,\AA{} cutoff on the
C$_\beta$ atoms for the Zhang benchmark, following
\citet{rao2021transformerplm} (the C$_\alpha$ atom substitutes for
glycine, which has no C$_\beta$), and an $8$\,\AA{} C$_\alpha$ cutoff for
CAMEO-PTA25. \citet{zhang2024cj} instead use a looser $10$\,\AA{}
C$_\alpha$ cutoff and report precision over the top $L$ pairs rather
than the top $L/2$, so our absolute precisions are not directly
comparable to the numbers in their paper; the comparison we rely on
throughout is the gap between methods under one fixed criterion.}
We restrict to
\emph{long-range} contacts ($|i-j| \geq 24$ along the chain), which are
the most informative for fold determination and the hardest to predict
from sequence alone \citep{dunn2008apc}. The contact map is the binary
symmetric matrix $C \in \{0,1\}^{L \times L}$ with $C[i,j]{=}1$ for
each long-range contact. Every method is evaluated by
\textbf{top-$L/2$ long-range precision} ($\text{P}@L/2$ long):
predict the $L/2$ highest-scoring long-range pairs, report the
fraction that are real contacts \citep{zhang2024cj}. All scores rank
from a single $(L,L)$ score map per method.

We use three evaluation sets, summarized in Table~\ref{tab:datasets}.
CAMEO-PTA25 is our primary, leakage-clean set: 553 single-chain CAMEO
targets deposited on or after 2022-09-01 (post-dating ESM-2's UniRef50
2021\_04 pretraining release), then filtered --- as a conservative proxy
for the pretraining-time release --- to \emph{retain only} targets with no
hit in current UniRef50 \citep{suzek2007uniref} at $\geq$50\% MMseqs2
\citep{steinegger2017mmseqs2} identity over $\geq$80\% query coverage,
following the leakage protocol of \citet{hermann2024leakage}; of the 56 targets
that pass, 39 retain $L \geq 75$ (the long-range eligibility cutoff), of
which 29 form the leakage-clean evaluation half. The \emph{PTA25} tag
marks the pretraining-aware CAMEO snapshot (assembled in 2025) rather
than asserting cleanliness outright, because leakage cleanliness is
model-relative. CAMEO is a moving target, and this split is leakage-clean
only for models whose pretraining predates the 2022-09 deposition floor
(e.g.\ ESM-2, ESM-1b, ProtT5). Models trained later, namely ESMC and
AMPLIFY (both 2024), may have seen targets in the post-2022-09 window, so
we treat them as \emph{partial} (AMPLIFY in
Table~\ref{tab:cameo-crossarch}; ESMC's within-model CAMEO margin in
Section~\ref{sec:fusion-beats-cj}).

\begin{table}[h]
\centering
\small
\caption{\textbf{Evaluation sets.} Selection budget is the disjoint
labeled subset on which the top-$K$ heads are picked; it never overlaps
the evaluation draw. The primary, leakage-clean comparison
(Section~\ref{sec:fusion-beats-cj}) runs on CAMEO-PTA25; cross-architecture
reproduction numbers (Appendix~Table~\ref{tab:a1}) and the methodology
analyses (operator comparison, $K$-sweep, cluster-diffuseness, depth
probe, timing) run on Zhang eval-200. CASP14-FM lists the 16 nominal
free-modeling targets; 14--16 are scored per model (a few long targets
are dropped at load on individual cells), with the exact per-cell $N$ in
Table~\ref{tab:a1}.}
\label{tab:datasets}
\resizebox{\textwidth}{!}{%
\begin{tabular}{lllcl}
\toprule
Dataset & Source & $N$ (eval) & Sel.\ budget & Role \\
\midrule
Zhang eval-200 & Zhang benchmark \citep{zhang2024cj} & 200 & select-10 & in-distribution repro. \\
CAMEO-PTA25 & CAMEO + Hermann filter & 29 & select-10 & leakage-clean primary \\
CASP14-FM & CASP14 \citep{kryshtafovych2021casp14} & 16 & --- & hard OOD \\
\bottomrule
\end{tabular}%
}
\end{table}

Zhang eval-200 is a 200-protein draw stratified by sequence length from
\citeauthor{zhang2024cj}'s 1{,}431-PDB-chain benchmark, with head and $K$
selection on a \emph{disjoint} 10-protein set (Zhang select-10) so
selection and evaluation never share a chain --- the in-distribution
analogue of the leakage-clean CAMEO split. CASP14-FM comprises 16
free-modeling targets of CASP14 \citep{kryshtafovych2021casp14} (mean
$L{=}423$, range 95--2180), a harder, longer test. The only 50-protein
Zhang draw (\emph{Zhang-50}) is the matched-budget supervised control in
Appendix~\ref{app:rao-baseline}, the larger budget the truncated-Rao
$L_1$-LR probe needs to fit; the selection-size study
(Appendix~\ref{app:selection-size}) uses the same pool and shows 10
labeled proteins suffice.

\subsection{Standard CJ and representation-CJ}
\label{sec:repr-cj-method}

\textbf{Standard CJ} \citep{zhang2024cj} --- which we also call
\textbf{logit-CJ}, because it reads the model's masked-LM logit output
--- builds a four-index Jacobian
$J[i, a', j, a] = \mathrm{logits}^{\,i\to a'}[j, a] - \mathrm{logits}_{\mathrm{unpert}}[j, a]$,
where the superscript $i\to a'$ denotes the forward pass on the sequence
with position $i$ substituted to amino acid $a'$, and
$\mathrm{logits}_{\mathrm{unpert}}$ is the native (unperturbed) forward
pass. The two amino-acid indices play distinct roles: $a'$ is the
\emph{substituted} amino acid placed at $i$, and $a$ is the
\emph{response} amino acid whose predicted logit is read at every other
position $j$. So $J[i,a',j,a]$ records how perturbing $i$ to $a'$ moves
the logit for $a$ at $j$, relative to the native sequence. Each position
is perturbed to its $19$ non-native amino acids, and one shared forward
pass supplies the native baseline --- so the per-protein cost is
$19L + 1$ forwards, not $20L$ (the native is the baseline, not a
perturbation). Standard preprocessing then collapses $J$ to an $(L, L)$
contact map: mean-center over the index axes, take the Frobenius norm
over the two amino-acid axes $a', a$, zero the diagonal,
apply average-product correction (APC; \citealp{dunn2008apc}, which
subtracts the per-row and per-column background co-variation that
otherwise swamps the true contact signal), and symmetrize.

Logit-CJ requires a per-position categorical head, so it applies only
to models with a compatible per-position categorical head (the ESM-family
models here; AMPLIFY lacks the exposed logit head logit-CJ needs in our
setup). We generalize
it to any architecture exposing per-position hidden states $h$ (the
residual-stream vector at a position and layer) by replacing the logit
response with a hidden-state response:
$J_{\mathrm{repr}}[i, a', j] = \| h^{\,i\to a'}[\ell, j] - h_{\mathrm{unpert}}[\ell, j] \|_2$,
the distance the hidden state at position $j$, layer $\ell$
(default = last; the layer sweep in Appendix~\ref{app:a4} confirms the
last layer is the right default) moves when $i$ is perturbed, in place of
how the logits move, averaged over $a'$ and given the same post-processing
as logit-CJ. We call this \textbf{repr-CJ}; it
applies to encoder-only, encoder-decoder, and causal models, at the
same $19L+1$-forward cost. On causal models we additionally replace
the symmetrize step with an upper-triangle-only readout
(\textbf{causal-CJ}). The two give an identical top-$L/2$ long-range
ranking: causal masking populates only one triangle of the response
(a position responds only to its left context), so symmetrizing merely
halves those entries relative to reading the populated triangle
directly --- a monotone rescaling that leaves the rank order of pairs,
and hence top-$L/2$ long-range precision, unchanged.

\paragraph{Why repr-CJ tracks logit-CJ: a metric-choice view.}
The two readouts measure the perturbation response in different spaces.
If the masked-LM head were a linear map $z = Wh$ from the final hidden
state $h \in \mathbb{R}^{d}$ to logits $z$, a hidden-state perturbation
$\Delta h$ would give $\Delta z = W\,\Delta h$ and
$\|\Delta z\|_2^2 = \Delta h^\top (W^\top W)\, \Delta h$: a logit readout
measures the displacement under the head-induced quadratic form
$W^\top W$ (positive-semidefinite, Mahalanobis-type), a hidden-state
readout under the Euclidean metric. This is the intuition for why the
two track each other, not an exact account of the headline numbers:
Standard/logit-CJ mean-centers a four-index tensor and takes a Frobenius
norm over the two amino-acid axes, whereas repr-CJ averages
per-substitution hidden-displacement norms, so the two differ in
centering and aggregation as well as in metric, and the per-protein
agreement (Pearson $r = 0.95$--$0.96$, Table~\ref{tab:repr-cj-arch})
reflects similar downstream behavior rather than the metric alone.

We isolate the metric in Appendix~\ref{app:wtw}: repr-CJ (Euclidean), a
$W^\top W$ readout (\textbf{WtW-CJ}, $\|W\Delta h\|_2$, $W$ the tied
output embedding), and a metric-matched logit readout are computed from a
single perturbation sweep with identical post-processing, so they differ
only in how the same hidden displacement is read out. Swapping Euclidean for
$W^\top W$ barely changes the readout (long-range pair Pearson $0.985$):
on the sampled contact-carrying displacements $W^\top W$ acts nearly like
a scaled Euclidean metric. This is near-isometry on those directions, not
global isotropy --- $W$ has at most $20$ amino-acid rows, so $W^\top W$
has rank $\leq 20$ (fewer after centering) and cannot be isotropic across
the $d$-dimensional hidden space; the agreement constrains it only on the
perturbations observed. Anisotropy is therefore one source of any
repr/logit divergence, alongside the head's nonlinearity and layer
normalization, the four-index centering, and finite (rather than
infinitesimal) substitutions, so we expect the readouts to come apart
when the head strongly distorts the relevant displacements. Because
repr-CJ swaps the learned output-head metric for Euclidean hidden-state
distance, it is best read as the \emph{output-head-independent} analogue
of logit-CJ rather than a literal metric variant; the full first-order
head metric is $J_f(h)^\top J_f(h)$ (the Jacobian of the whole head), of
which $W^\top W$ is only the tied-embedding term.

\subsection{Top-$K$ attention fusion}
\label{sec:fusion}

For each (layer, head) of an attention-LM we extract the
per-head attention map (symmetrize, APC, zero-diagonal; for encoder--decoder
models this is the encoder self-attention) and score it
by $\text{P}@L/2$ long against a small labeled selection set, disjoint
from the evaluation set (both the Zhang eval-200 and the CAMEO-PTA25
results select on a disjoint 10-protein subset; only the matched-budget
control against the $L_1$-LR in Appendix~\ref{app:rao-baseline} uses 50,
the larger budget needed to fit that regression). A selection-set-size sweep
(Appendix~\ref{app:selection-size}) shows 10 proteins suffice.
Averaging precisions (unweighted over selection proteins) gives a global head ranking; the \textbf{top-$K$
heads} are the only labeled step. Our default operator is the
arithmetic mean over the top-$K$ APC'd maps:
$S = \tfrac{1}{K} \sum_{k=1}^{K} A_{k}$, scored by
$\text{P}@L/2$ long. Per-test cost: \textbf{one forward of the
length-$L$ sequence}.

We sweep $K \in \{1,2,3,5,7,10,20\}$ (Section~\ref{sec:ksweep}) and
ablate the arithmetic mean against nine alternative operators
(elementwise max, geometric mean, z-score mean, reciprocal-rank
fusion, positional-filter mean, entropy-gated mean, graph closure,
modern-Hopfield, spectral consensus; Appendix~\ref{app:operators}).
The arithmetic mean is the per-cell best on 8 of 16 cells and within
$0.3$pp of the best alternative on most others; we adopt it as the
cross-architecture default for simplicity. An APC ablation
(Appendix~\ref{app:apc-ablation}) confirms the fusion gain over the
top-1 head comes from averaging, not from APC.

\subsection{Models}

We evaluate nine protein-LM variants spanning masked-LM (ESM-2 at
five scales --- 8M / 35M / 150M / 650M / 3B; ESM-1b; AMPLIFY-350M),
span-corruption encoder--decoder (ProtT5-XL), and autoregressive
causal-LM (ProGen2-xlarge) objectives. Per-model parameter counts,
position encodings, pretraining corpora, and checkpoint identifiers are
in Appendix~\ref{app:confounds}--\ref{app:artifacts}. Whether CJ's
contact-prediction result is a property of the Jacobian readout or of
ESM-2 specifically was not addressed in the original work
\citep{zhang2024cj}. The Categorical Jacobian has since been applied to
other masked-LM and genomic sequence models (e.g.\ ESM3, Evo, gLM2),
but, to our knowledge, not to an encoder--decoder or a causal protein
language model; the evaluation below is, to our knowledge, the first
such cross-objective test. Significance is assessed by paired bootstrap on
per-protein $\text{P}@L/2$ long deltas (Appendix~\ref{app:bootstrap});
code, configurations, and the summary result tables are released with
the paper.

\section{Results}

\subsection{Fusion reads contacts more confidently than perturbation}
\label{sec:confidence}

\begin{figure}[t]
  \centering
  \includegraphics[width=\linewidth]{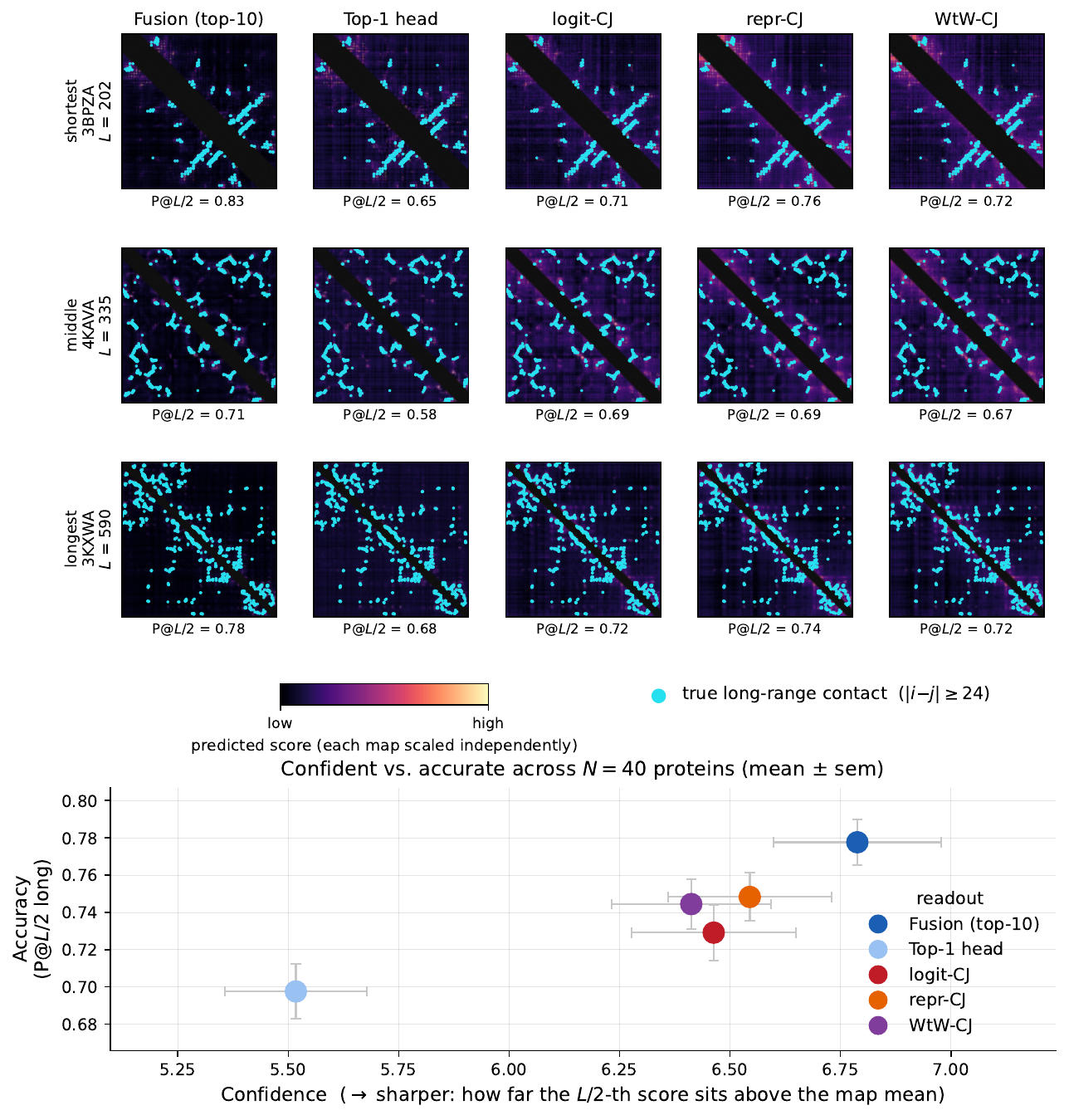}
  \caption{\textbf{Fusion is the most confident contact readout.} \emph{Top:}
  per-method score maps (long-range region; magma, short-range band $|i-j|<24$
  masked) for the shortest, middle, and longest proteins of a length-stratified
  ESM-2-650M Zhang eval-200 sample; true long-range contacts overlaid in cyan,
  with per-cell top-$L/2$ long precision. Top-$K$ attention fusion concentrates
  its score mass on a few pairs that track the true contacts; the Categorical
  Jacobian and its representation variants (logit-/repr-/WtW-CJ) spread the same
  signal more diffusely, and the single best head is noisier still.
  \emph{Bottom:} across $N{=}40$ proteins, fusion alone occupies the
  high-confidence, high-precision corner --- it is both the sharpest
  readout (its $L/2$ decision-cutoff score sits furthest above the map mean) and
  the most accurate. Metric definitions and the per-method table are in
  Appendix~\ref{app:confidence}.}
  \label{fig:confidence}
\end{figure}

Before the quantitative comparisons, Figure~\ref{fig:confidence} shows what the
methods actually produce. We read a readout's \emph{confidence} off how
decisively its score map concentrates on a few predicted contacts, its
sharpness: how concentrated the score mass is (Gini) and how far the
top-$L/2$ decision cutoff sits above the map mean (a $z$-score). Alongside it we
report two discriminability diagnostics that are not confidence measures:
a $d'$ label separation (the gap between the mean true-contact and
non-contact scores, in units of their pooled standard deviation) and
top-decile enrichment, to check whether sharpness coincides with better
discrimination (Appendix~\ref{app:confidence}). On a length-stratified
ESM-2-650M Zhang eval-200 sample, top-$K$ fusion is the most confident readout:
its score mass is by far the most concentrated (Gini $0.26$ vs.\
$0.12$--$0.17$ for the perturbation methods and the single head, at the highest
cutoff $z$), and that confidence is warranted, because fusion is also the most
accurate (P@$L/2$ long $0.78$), the only method in the high-sharpness,
high-precision corner of Figure~\ref{fig:confidence} (bottom). On the
label-separation diagnostic the perturbation readouts are
marginally higher ($d'$: repr-CJ $1.35$ vs.\ fusion $1.24$), so fusion's
edge is concentration (it commits to a few pairs and is right) rather than
raw discriminability, which the perturbation methods spread more evenly across
the map. The single best head is both the least concentrated and the least
accurate.

\subsection{Fusion of top-$K$ attention heads beats CJ across architectures}
\label{sec:fusion-beats-cj}

Any single-set claim involving PLMs is open to the criticism that the model
has memorized the evaluation set during pretraining \citep{hermann2024leakage}.
We therefore lead with a leakage-clean readout, treat the pretraining-aware
CAMEO set as the primary evaluation, and use Zhang eval-200 \citep{zhang2024cj} to
reproduce numbers on a comparable benchmark.

\paragraph{Primary result: end-to-end leakage-clean.}
The comparison the paper rests on removes pretraining leakage end to end.
On the pretraining-aware CAMEO-PTA25 split (Section~\ref{sec:setup}: 56 of
553 post-cutoff targets survive the \citet{hermann2024leakage} leakage
filter, only $\sim$10\%, a measure of how leakage-prone contemporary CAMEO
is, and 39 retain $L \geq 75$; construction and the conservative
current-UniRef50 substitution in Appendix~\ref{app:uniref50}), neither head
selection nor evaluation touches data the models have plausibly memorized.

The cleanest comparison runs head selection and evaluation on
\emph{disjoint halves} of CAMEO-PTA25, so no Zhang chains appear
anywhere in the pipeline. We select the top-$K$ heads on a
length-stratified 10-protein subset of the $L \geq 75$ CAMEO-PTA25
targets and evaluate fusion on the disjoint remaining $L \geq 75$
targets ($N{=}29$). This uses a 10-protein selection budget (matching the
in-distribution protocol) rather than the larger half-split, which
leaves more targets for evaluation. On ESM-2-650M, fusion beats Standard
CJ significantly (Table~\ref{tab:cameo-clean}); even the globally selected
top-1 head already matches CJ ($0.396$ vs $0.369$, n.s.), so on
leakage-clean data CJ's $\sim$$19L$ perturbations buy nothing over
reading a single head; fusion ($0.459$) then beats CJ by $+9$pp and
the top-1 head by $+6$pp. We fix
$K{=}10$ (the cross-architecture default).

\begin{table}[h]
\centering
\small
\caption{\textbf{End-to-end leakage-clean comparison on CAMEO-PTA25
($L \geq 75$, $N {=} 29$).} Top-$K$ heads selected on a disjoint
10-protein subset; per-protein paired bootstrap, $B{=}10{,}000$.
Asterisks: $^{***}p<0.001$, $^{**}p<0.01$, $^*p<0.05$.}
\label{tab:cameo-clean}
\begin{tabular}{lcc}
\toprule
Method (ESM-2-650M) & $\text{P}@L/2$ long & $\Delta$ vs CJ \\
\midrule
Per-protein top-1 head & 0.396 & $+0.027$ (n.s.) \\
Standard CJ            & 0.369 & --- \\
\textbf{Fusion ($K{=}10$, naive-mean)} & \textbf{0.459}$^{***}$ & $\mathbf{+0.090^{***}}$ \\
\bottomrule
\end{tabular}
\end{table}

\paragraph{Pretraining-aware cross-architecture.}
Running the same end-to-end leakage-clean protocol --- head selection on a
disjoint CAMEO select-10 subset, evaluation on the remaining $N{=}29$
targets, so no Zhang chains touch the pipeline --- reproduces the pattern
across architectures (Table~\ref{tab:cameo-crossarch}): fusion beats CJ by
$+9.0$, $+9.4$, $+5.9$, and $+5.9$pp on ESM-2-650M, ESM-2-3B, ProtT5-XL,
and ESM-1b (all $p{<}0.001$, paired bootstrap over $N{=}29$). These four
comparisons read the \emph{same} 29 evaluation proteins with different
models, so they are correlated reads of one small draw rather than four
independent replications; the consistent within-model sign of the margin
is the robust signal, not four separate significance tests. These clean
margins are \emph{larger} than the partial-clean operating point we
reported previously (selection on a contaminated corpus gave
$+7.2$/$+7.4$/$+5.2$pp): clean selection removes contamination that had
inflated CJ, widening the gap. For AMPLIFY-350M, which has no masked-LM
logit head, we report fusion vs.\ top-1 only ($+0.65$pp); ESM-1b's
fusion-vs-top-1 margin is \mbox{$+0.8$pp}. On these clean data, ESM-2-3B
edges ESM-2-650M only marginally ($0.463$ vs.\ $0.459$), far below the
in-distribution scaling gap (3B leads 650M by $\sim$2.8pp on Zhang
eval-200): the larger model's in-distribution edge is partly the
memorization the Hermann filter removes.

\begin{table}[t]
\centering
\small
\caption{\textbf{CAMEO-PTA25 cross-architecture comparison
(end-to-end leakage-clean).}
$\text{P}@L/2$ long on CAMEO-PTA25 ($L \geq 75$); top-$K$ heads
selected on a disjoint length-stratified 10-protein subset (CAMEO
select-10) and evaluated on the remaining $N{=}29$ targets, so no
Zhang chains appear anywhere in the pipeline. $K{=}10$ throughout
(the cross-architecture default). Asterisks on $\Delta$ are
paired-bootstrap significance ($^{***}p{<}0.001$, $^{**}p{<}0.01$,
$^*p{<}0.05$). AMPLIFY-350M shows fusion vs.\ top-1 head only (it
has no masked-LM logit head for logit-CJ); ProGen2-xlarge is
treated as the causal-LM scope boundary in
Section~\ref{sec:causal}. AMPLIFY-350M is marked \emph{partial} for
CAMEO-PTA25 leakage cleanliness (Appendix~\ref{app:confounds}).}
\label{tab:cameo-crossarch}
\begin{tabular}{llcccc}
\toprule
Architecture & $K$ & top-1 head & fusion & CJ & $\Delta$(fusion -- CJ) \\
\midrule
ESM-2-650M    & 10 & 0.396 & \textbf{0.459} & 0.369 (logit) & $+0.090^{***}$ \\
ESM-2-3B      & 10 & 0.347 & \textbf{0.463} & 0.369 (logit) & $+0.094^{***}$ \\
ProtT5-XL     & 10 & 0.354 & \textbf{0.418} & 0.359 (repr)  & $+0.059^{***}$ \\
ESM-1b        & 10 & 0.340 & \textbf{0.348} & 0.289 (logit) & $+0.059^{***}$ \\
AMPLIFY-350M$^{\dagger}$  & 10 & 0.255 & \textbf{0.262} & ---           & ---            \\
\bottomrule
\multicolumn{6}{l}{\footnotesize $^{\dagger}$ partial leakage cleanliness: AMPLIFY pretraining post-dates the} \\
\multicolumn{6}{l}{\footnotesize \phantom{$^{\dagger}$ }2022-09 CAMEO deposition cutoff, so its absolute precision is not leakage-clean;} \\
\multicolumn{6}{l}{\footnotesize \phantom{$^{\dagger}$ }the within-model $\Delta$(fusion\,--\,CJ) largely controls for it (both read the same model).} \\
\end{tabular}
\end{table}

\paragraph{Zhang eval-200 reproduction and length scaling.}
On the in-distribution Zhang eval-200 (direct reproduction comparability
with \citet{zhang2024cj}), fusion beats CJ on every architecture at and
above the 35M scale with paired-bootstrap significance at $p<0.01$
(Figure~\ref{fig:headline}; per-cell numbers in
Appendix~Table~\ref{tab:a1}). Two in-distribution exceptions. On the
smallest ESM-2-8M the $K{=}10$ default over-selects a
concentrated-cluster model (best read at $K{=}3$,
Section~\ref{sec:ksweep}) and fusion trails CJ ($0.211$ vs $0.234$). On
the later-added ESMC, fusion and logit-CJ are within noise
($+0.004$ on $300$M; the gap opens to a significant $+0.06$--$0.08$
only on the harder leakage-clean CAMEO split, Table~\ref{tab:cameo-crossarch}). On the harder, longer CASP14-FM
(mean $L{=}423$) the gap widens, and \emph{fusion on the smaller
650M exceeds CJ on the larger 3B} --- a length-scaling pattern
consistent with, though not proving, greater length sensitivity in
CJ's perturbation-based readout relative to fusion's single forward.
Per-architecture top-$K$ values identified on the selection
set transfer to a disjoint held-out draw without re-tuning
(Appendix~\ref{app:a5}).

For absolute calibration: ESMFold's PAE contacts \citep{lin2023esm2}
reach $\text{P}@L/2$ long $\approx 0.6$--$0.8$ on contemporary
contact-prediction benchmarks; supervised MSA-augmented attention
readouts \citep{rao2021msa} reach $\approx 0.6$--$0.7$. Our
leakage-clean operating point ($\approx 0.4$ absolute) sits below both,
as expected for a label-efficient, MSA-free method on data the
model has not memorized. The interpretable comparison is the
$+5$-pp range gap over CJ.

\paragraph{A recent open backbone: ESM Cambrian.}
The central result holds on a recent open architecture: fusion beats
the Categorical Jacobian on ESMC, and its within-model margin matches
what ESM-2 showed two model generations earlier. ESMC
\citep{esmteam2024esmc,biohub2026esm} is the bidirectional masked-LM
successor to ESM-2, and we apply logit-CJ and fusion to it without
modification. On in-distribution Zhang eval-200, ESMC's higher contact
scores at smaller parameter counts coincide with its stronger backbone:
logit-CJ exceeds the same-tier ESM-2 baseline (both ESMC bootstrap CIs
lie above the ESM-2 point estimate) and fusion roughly matches it, at
$2$--$5\times$ fewer parameters (per-tier numbers in
Table~\ref{tab:repr-cj-arch}).

These absolute gains carry a bound. ESMC's 2024 pretraining post-dates
both Zhang eval-200 and our 2022-09-cutoff CAMEO-PTA25 split, so neither
is leakage-clean for ESMC, as is the case for the post-cutoff AMPLIFY.
The leakage-robust signal is the \emph{within-model} fusion-over-CJ
margin, where both methods read the same (possibly memorized) model: on
CAMEO-PTA25 ESMC's paired $\Delta$(fusion\,--\,CJ) is $+0.063$ ($300$M)
and $+0.078$ ($600$M), both $p<0.001$, comparable to ESM-2's margin on
the matched partial-clean protocol ($+0.072$ / $+0.074$). On the fully
leakage-clean select-10/eval-29 split ESM-2's margin widens to
$+0.090$ / $+0.094$ (Table~\ref{tab:cameo-crossarch}).
A post-2024 deposition split would be needed to confirm ESMC's
absolute advantage on fully leakage-clean data.

\subsection{Representation-CJ generalizes Standard CJ to non-MLM architectures}
\label{sec:repr-cj}

Logit-CJ requires a per-position categorical head, restricting it to
bidirectional masked-LM architectures. Repr-CJ (Section~\ref{sec:repr-cj-method})
substitutes a per-position hidden-state response for the logit
response, which extends the Jacobian construction to
encoder--decoder and causal architectures at the same
$19L{+}1$-forward cost. We validate repr-CJ against logit-CJ on the
four MLM cells where both apply (per-cell Pearson $r$ in
Table~\ref{tab:repr-cj-arch}): the per-protein scores are tightly
correlated on both ESM-2 scales at Zhang eval-200, where the logit-CJ and
repr-CJ means also agree within bootstrap noise (the eval-200 scatter is
in Appendix~Figure~\ref{fig:repr-cj}), and the two ESMC scales replicate
the tight correlation on Zhang-50. Repr-CJ thus extends the Jacobian
construction faithfully where logit-CJ is undefined; per-architecture
numbers are in
Table~\ref{tab:repr-cj-arch}. Beyond aggregate precision, the two readouts
also agree at the \emph{pair} level (top-$L/2$ long-range Jaccard $0.73$,
long-pair Pearson $0.95$ on ESM-2-650M), and a $W^\top W$-metric readout
inserted between them supports the metric-choice account of
Section~\ref{sec:repr-cj-method} (Appendix~\ref{app:wtw}).

\begin{table}[h]
\centering
\small
\caption{\textbf{Representation-CJ across architectures (Zhang eval-200,
last-layer hidden states).} Repr-CJ tracks logit-CJ within bootstrap
noise on every cell where both are defined. ProtT5-XL's repr-CJ is
competitive with MLM models of similar size and exceeds its per-protein
top-1 attention head. AMPLIFY-350M's repr-CJ underperforms its top-1
head; a layer sweep (Appendix~\ref{app:a4}) confirms the last layer is
the correct readout, so the gap is intrinsic to AMPLIFY. The ESMC
repr-CJ entries ($\ddagger$) are Zhang-50 values, not a like-for-like
eval-200 comparison.\protect\footnotemark}
\label{tab:repr-cj-arch}
\begin{tabular}{lcccc}
\toprule
Architecture & top-1 head & logit-CJ & repr-CJ & Pearson $r$ \\
\midrule
ESM-2-650M  & 0.678 & 0.733 & 0.732 & 0.96 \\
ESM-2-3B    & 0.722 & 0.742 & 0.746 & 0.95 \\
ESMC-300M   & 0.721 & 0.760 & 0.768$^{\ddagger}$ & 0.92$^{\ddagger}$ \\
ESMC-600M   & 0.747 & 0.787 & 0.790$^{\ddagger}$ & 0.90$^{\ddagger}$ \\
ProtT5-XL   & 0.644 & ---   & 0.700 & ---  \\
AMPLIFY-350M& 0.510 & ---   & 0.490 & ---  \\
\bottomrule
\end{tabular}
\end{table}
\footnotetext{$^{\ddagger}$ESMC repr-CJ and its Pearson $r$ are the
Zhang-50 values; the eval-200 rerun is pending because the Biohub-fork
ESMC attention is unfused, making the $\sim 19L$ repr-CJ perturbation
forwards impractically slow (ESMC logit-CJ and fusion are at eval-200).}

\subsection{How contact signal is distributed across attention: diffuseness predicts $K$}
\label{sec:ksweep}

Sweeping $K \in \{1, 2, 3, 5, 7, 10, 20\}$ on Zhang eval-200 (full numbers in
Appendix Figure~\ref{fig:ksweep}), four of the nine variants peak at
$K \in \{2, 3, 7\}$; pushing $K$ higher costs precision on the
concentrated-cluster architectures (e.g.\ AMPLIFY-350M drops $\sim$1.7pp
from its $K{=}3$ peak when extended to $K{=}10$, and $\sim$6.9pp by $K{=}20$).
A uniform $K{=}10$ default is not the right choice across the architecture set.

To understand the variation, we measure \textbf{head-cluster
diffuseness}: the fraction of proteins in the selection set for which
the globally top-ranked head (ranked by mean P@$L/2$ long across the
selection set) is also that protein's individually best head
(Appendix Figure~\ref{fig:cluster}). This single statistic cleanly
separates two regimes. In \emph{concentrated-cluster} models
($>65\%$ top-1 win rate) --- ESM-2-8M, AMPLIFY-350M, ProGen2-xlarge
--- one dominant head carries the contact signal for most proteins,
and averaging in additional heads dilutes signal with noise; optimal
$K \leq 3$. In \emph{diffuse-cluster} models ($<50\%$ win rate) ---
ESM-2-650M, ESM-1b, ProtT5-XL --- different heads carry different
proteins' contact signal, and no single head dominates; averaging
recovers a consensus that no individual head expresses on its own,
and optimal $K \in \{7, 10\}$.

Why do models differ? One possibility is that models with more
capacity (more heads, more layers) can afford to spread pair
structure across a larger pool of heads rather than concentrating
it in one, because each head can specialize on a subset of
structural motifs while the ensemble recovers the full contact map.
Consistent with this reading, the concentrated/diffuse split is not
a property of model family or training objective alone: it varies
within the ESM-2 family itself (8M concentrated, 650M diffuse, 3B
intermediate). Different PLMs --- even at the same pretraining
objective and within the same training pipeline --- learn
substantively different functions on their attention heads. The shape
of the head-precision rank distribution signals the right $K$:
concentrated clusters (a sharp elbow) favor small $K$, diffuse clusters
(a long tail) favor large $K$ (Appendix~\ref{app:b}).

\citet{bhattacharya2021attention} reported that contact-relevant heads in
ProtBERT-BFD cluster in the late layers. The same late-layer concentration
holds on ESM-2-650M (top-$K$ on $\ell \in \{28\text{--}32\}$ recovers
$\sim$1pp below global top-$K$); restricting head selection to a late-layer
band is approximately right but not strictly so --- two contact-relevant
heads at $\ell{=}26,27$ on ESM-2-650M fall outside any reasonable late-layer
window. Global selection over all layers is the more conservative recipe.

\subsection{Causal language models: a scope boundary}
\label{sec:causal}

Causal masking enforces $h_j = f(s_{\leq j})$: each position attends
only to its left context. We hypothesize this leaves less room for the
symmetric pair statistic that attention-readout methods exploit ---
though directional attention can in principle still feed a symmetric
downstream readout. We test it on two
ProGen2 variants \citep{madani2023progen2}: ProGen2-xlarge ($\sim$6.4B
params) and ProGen2-large ($\sim$2.7B params), both trained
autoregressively on UniRef90. Every attention-readout
collapses on both variants: top-1 head reaches $0.193$ / $0.033$
(xlarge on Zhang eval-200, vs.\ $0.678$ on ESM-2-650M; large on Zhang-50),
repr-CJ collapses to $0.056$ / $0.013$, and fusion to $0.112$ / $0.023$. The head-precision profile is flat on both
(Appendix~\ref{app:a2}): no head shows the precision spike of the MLM
contact-head cluster \citep{bhattacharya2021attention}, and the
smaller variant is worse, not better --- ruling out a
scale-specific artifact.

\subsection{Compute economy}
\label{sec:cost}

\begin{figure}[t]
  \centering
  \includegraphics[width=\linewidth]{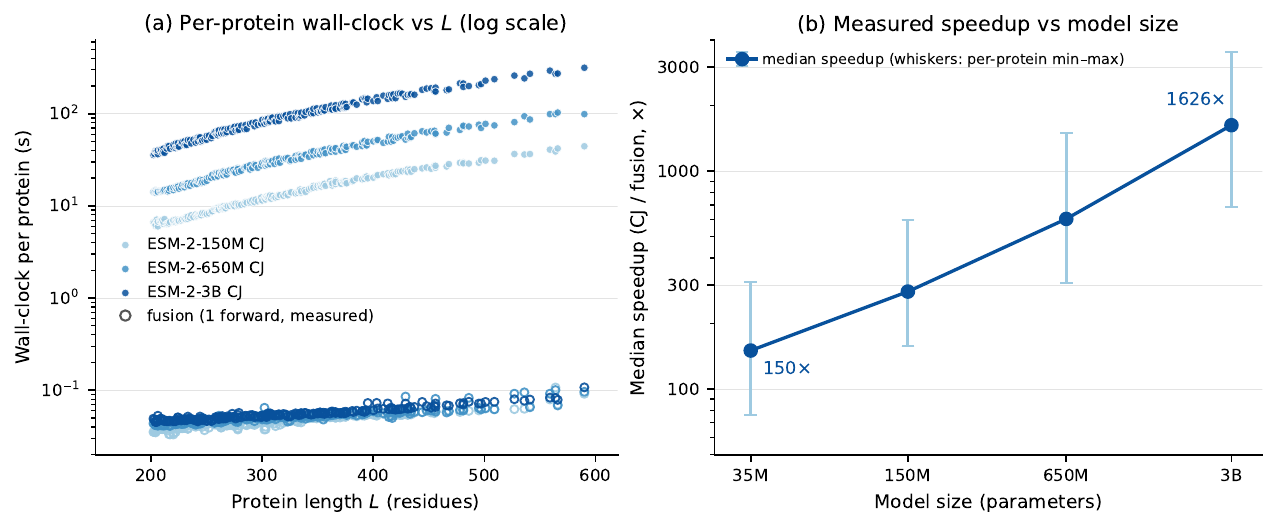}
  \caption{\textbf{Compute economy.} (a)~Per-protein wall-clock vs.\
  protein length $L$ on ESM-2-\{150M, 650M, 3B\} (log $y$): CJ (filled)
  scatters along the $\sim 19L$-forward trend; fusion (open markers) is a
  real per-protein measurement on the \emph{same} proteins and hardware
  --- it grows mildly with $L$ but sits $2$--$3$ orders of magnitude below
  CJ. The gap is the $\sim 19L$-vs-$1$ forward-count ratio, not
  length-independence.
  (b)~Measured per-protein wall-clock speedup (CJ\,/\,fusion) vs.\ model
  size on ESM-2-\{35M, 150M, 650M, 3B\} (Zhang eval-200, log--log; markers
  are the per-protein median, whiskers the per-protein min--max): the
  advantage grows monotonically with scale, from $\sim 150\times$ at 35M to
  $\sim 1600\times$ at 3B, because CJ's $\sim 19L$ forwards cost more on a
  larger backbone while fusion stays at one. Every point is measured on
  identical A100-80GB hardware (Table~\ref{tab:a5}), not extrapolated.}
  \label{fig:cost}
\end{figure}

The cost difference between CJ and fusion is structural, not
incidental. CJ must \emph{construct} a pairwise score by varying the
input: for each of the $L$ residue positions, it substitutes each of
the 19 alternative amino acids and runs a forward pass to measure how
every other position's masked-LM logits respond. The cost is
$19 \times L$ perturbation forwards per protein --- 19 substitutions at
each of the $L$ positions --- where the 19 forwards at a given position
$i$ are aggregated (Frobenius norm over the response amino acids) into a
single column of the $(L \times L)$ Jacobian, so the $L$ positions yield
its $L$ columns. These perturbations cannot be
collapsed into a single forward pass because each one changes the
input sequence at a different position. Fusion, by contrast, reads
the $(L \times L)$ attention maps that the model already computes in
its native forward pass --- the pair structure is a free
byproduct of how the transformer processes the sequence, not
something that needs to be reconstructed via perturbation. The
per-protein cost of fusion is therefore one forward pass at test
time, plus a one-time head-probe cost on a small labeled set (10
proteins) that is shared across all subsequent test proteins.
Figure~\ref{fig:cost} summarizes the per-protein wall-clock and the
measured per-model speedup; the underlying per-protein timing table is in
Appendix~\ref{app:a5} (Table~\ref{tab:a5}). Fusion's one-time
10-protein head probe (one per model) is only $\sim$10 forward passes --- negligible
against CJ's $\sim 19L$ forwards per test protein --- so fusion is
cheaper from small $N$; the gap widens
with both
model size and protein length because CJ needs $\sim 19L$ forward
passes to fusion's one. Fusion's number of model evaluations does
not grow with $L$; its wall-clock still does --- a single forward is at
least $O(L^2)$ in the attention --- but far more slowly than CJ's
$\sim 19L$ forwards. The measured per-protein advantage is therefore
large and grows with model size: on identical A100-80GB hardware and
precision, median wall-clock speedups run from $\sim 150\times$
(ESM-2-35M) to $\sim 1600\times$ (ESM-2-3B) on Zhang eval-200, and
$\sim 466\times$ on the CASP14-FM targets both methods can
run (Appendix~\ref{app:a5}); fusion's peak GPU memory is also
$2$--$5\times$ lower. This is measured against the batched CJ
implementation (perturbations batched for memory, $\sim$32--64 per
forward), not an unbatched strawman.

Fusion also scales past the backbone's context window. A protein
longer than the model's context can be read either by tiling it into
overlapping windows and stitching the per-window maps, or by using a
backbone with a longer or extrapolating context (ProtT5's relative-position
attention; ESMC's $2{,}048$-token rotary context). Stitching is
nearly lossless on feasible proteins, and on the two CASP14-FM targets that
exceed ESM-2's $1{,}024$-token limit both routes recover comparable contacts
at $\sim$1\,s of forwards --- where a tiled Categorical Jacobian would cost
$\sim 19W$ per window. We develop this in Appendix~\ref{app:stitch}.

\section{Discussion}
\label{sec:discussion}

\paragraph{What the gap means.}
The Categorical Jacobian reads contact signal indirectly via
$\sim 19L$ perturbations of the masked-LM head; fusion reads the same
signal off attention directly in one forward pass. That fusion beats
CJ on leakage-clean data across every comparable architecture is
evidence the model already exposes the structural pair information CJ
reconstructs. Three corollaries. \emph{(i)} CJ does not robustly beat
the best single attention head on leakage-clean data
(Section~\ref{sec:fusion-beats-cj}); its Zhang eval-200 (in-distribution) edge over the
top-1 head is partly a memorization edge. \emph{(ii)} A supervised
$L_1$-LR at the same 50-protein label budget reaches the same
precision as the unweighted top-$K$ mean on the clean control
(Appendix~\ref{app:rao-baseline}): the operative variable is the
labeled head selection, not the parametric form of the readout.
\emph{(iii)} The 30--36pp absolute drop from Zhang eval-200 to
CAMEO-PTA25 (fusion $\sim$30pp, CJ $\sim$36pp) is observed for both
methods --- the relative ordering of methods survives the filter, the
absolute numbers do not, and the gap upper-bounds how much of prior
unsupervised contact-prediction precision reflected pretraining overlap
(the CAMEO difficulty shift contributes the remainder).

\paragraph{Diffuseness as a property of learned attention.}
The concentrated-vs-diffuse split (Section~\ref{sec:ksweep}) is not
predicted by model family, scale, or training objective alone --- it
varies within the ESM-2 family itself. Different PLMs trained under the
same objective allocate pair structure across attention heads in
substantively different ways: some concentrate it on a single dominant
head, others spread it across a cluster. The phenomenon is consistent
with mechanistic-interpretability work on PLM circuits
\citep{adams2025plmsae,simon2025interplm} that finds discrete,
identifiable features carrying structural information. A useful
practical consequence is that a small head-probe diagnostic
classifies a new architecture into its diffuseness regime
(Appendix~\ref{app:b}) and picks $K$ by a short sweep accordingly.

\paragraph{The causal-LM boundary.}
Causal masking ties each position's representation to its left context.
We read the collapse as consistent with bidirectional pretraining
mattering for the symmetric pair statistic that attention-readout
methods exploit, rather than as an architectural impossibility ---
directional attention could still support a symmetric readout in
principle. Both ProGen2 variants show flat
head-precision profiles and complete collapse of fusion and repr-CJ
(Section~\ref{sec:causal}); the smaller variant is worse, not better,
ruling out a scale-specific artifact. The boundary now holds on two
causal PLMs within one family; a cross-family test (different causal
PLM, different training corpus) or an MLM ablation of the ProGen2
backbone would tighten it further.

\paragraph{Where in the network the contact signal lives.}
The top-$K$ heads we read live in the late layers of ESM-2-650M
(Section~\ref{sec:ksweep}). Is the signal generated late, or
does the late-layer readout sit on top of pair structure built up over
many layers? A sustained mean-ablation depth probe answers the second
(Appendix~\ref{app:depth-probe}): replacing the residual at layer
$\ell \in \{16, 24, 28, 32\}$ with the corpus-mean residual and
propagating forward, CJ precision decays monotonically with depth ---
no localized drop at any single layer --- which says the late-layer
readout sits on top of pair structure the network has been assembling
since early-to-mid depth. This is consistent with circuit-analysis findings
that PLM features compose through depth
\citep{adams2025plmsae,simon2025interplm}.

\paragraph{Limitations.}
The fusion method needs a small labeled head-selection set ($10$ proteins
in our in-distribution evaluation, $50$ for the matched-budget $L_1$-LR
control), a labeled step CJ does not require; we mitigate by showing $K$ transfers
cleanly to a disjoint sample (Appendix~\ref{app:a5}) and by
demonstrating that as few as 10--20 labeled proteins recover most of
the head ranking (Appendix~\ref{app:selection-size}). Our
leakage-clean evaluation is a sequence-similarity filter at 50\%
identity, which controls for direct sequence overlap between
pretraining and evaluation but not for fold-level overlap; a stricter
structural-similarity filter against the pre-cutoff PDB would
tighten the lower bound on memorization (we leave it to future work).
We do not report results on the multi-track ESM-3 family
\citep{hayes2025esm3} or on a causal PLM outside the ProGen2 family,
and we test only protein language models: because fusion reads native
attention, the same readout should extend to any sequence language model
with self-attention, and the Categorical Jacobian already runs on genomic
LMs (Evo, gLM2), so nucleic-acid contact readout is a natural extension.
Both methods also inherit the backbone's context-window bound: ESM-2's
$1{,}024$-token limit means neither the single fusion forward nor CJ's
$\sim 19L$ sweep can process a longer chain in one piece (two CASP14-FM
targets, $L{=}1372$ and $2180$, are excluded from every timing and
accuracy number above). Longer chains can be reached either by tiling
into overlapping windows and stitching a short-context backbone, or by
running a longer-context backbone natively (ProtT5's relative-position
attention, ESMC's $2{,}048$-token rotary context; neither truly unbounded,
Appendix~\ref{app:stitch}): stitching is nearly lossless on
feasible proteins, and both routes recover comparable contacts on two real
$L{>}1024$ targets in $\sim$1\,s of forwards, where a tiled Categorical
Jacobian is prohibitive. A full long-protein benchmark remains future work.
For ESMC, both Zhang eval-200 and our 2022-09-cutoff CAMEO-PTA25 split
pre-date its 2024 pretraining, so the in-distribution comparisons
ride backbone quality but are not leakage-clean for ESMC; the
leakage-robust observation is the within-model fusion-over-CJ margin,
which a post-2024 deposition split would let us confirm absolutely.
Finally, ensembling CJ with the head readout (per-protein z-score, then
sum) does not beat the head readout alone --- the gain is flat across
three cells and slightly negative on ESM-2-3B, the cell where fusion
leads CJ most. The two methods do disagree at the per-pair level
(top-$L/2$ Jaccard $\approx 0.59$), but that disagreement does not
convert into a robust precision gain, so we find no evidence CJ carries
contact signal the selected heads lack.

\section{Conclusion}

The central finding of this paper is that pre-trained protein language
models already expose long-range contact structure directly through
their attention maps. A small, identifiable cluster of attention heads
carries this signal, and the naive arithmetic mean of that cluster ---
with no learned weights at inference time --- is a sharper label-efficient
contact predictor than the Categorical Jacobian's perturbation-based
readout, at a fraction of the compute. This suggests that the
Jacobian's $\sim 19L$ perturbations are not \emph{extracting} contact
information so much as reconstructing what the model already
represents in its attention patterns.

The leakage-clean evaluation reveals a second finding: both methods drop
30--36 percentage points absolute from in-distribution to
leakage-clean evaluation, which suggests prior unsupervised
contact-prediction benchmarks have been overestimating both methods'
true generalization --- though the CAMEO-vs-Zhang difficulty shift
contributes to the gap, so we read it as an upper bound on the leakage
component rather than a measurement of it. Future work in this area
should adopt pretraining-aware evaluation as a default, not a
sensitivity analysis.

Finally, how the contact-head cluster is distributed across attention
heads varies by architecture and predicts the right $K$ for fusion.
This diffuseness pattern is not an artifact of model family or scale
--- it varies within the ESM-2 family itself --- which suggests that
PLM contact-encoding is a learned allocation property of the
pretraining process rather than an architectural inevitability. The
collapse of both fusion and CJ on the causal language models we tested
is consistent with bidirectional pretraining being important, though with
two same-family causal models we report this as associational
(see Section~\ref{sec:causal}, Appendix~\ref{app:confounds}).

\section*{Acknowledgments}

We thank Marmik Chaudhari for the discussions which inspired this work.
We thank the maintainers of the ESM-2, ESM-1b, AMPLIFY, ProtT5, and
ProGen2 model checkpoints, the UniRef and CAMEO consortia, and the
MMseqs2 development team for their open releases of the artifacts
this work depends on.

\section*{Use of large language models}

Large language models were used throughout this work --- for code,
data analysis, and drafting --- under the authors' direction and review.
The authors take full responsibility for all claims, results, and
content.

\section*{Conflicts of interest}

The authors declare no competing interests.

\bibliographystyle{plainnat}
\bibliography{references}

\appendix
\renewcommand{\thefigure}{A\arabic{figure}}
\renewcommand{\thetable}{A\arabic{table}}
\setcounter{figure}{0}
\setcounter{table}{0}

\section{UniRef50 release choice for the Hermann filter}
\label{app:uniref50}

The Hermann \citep{hermann2024leakage} filter requires that each
test sequence have no high-identity hit against a database
representative of the pretraining corpus. ESM-2 was pretrained on the
UniRef50 2021\_04 release \citep{lin2023esm2}; we filter against the
current (2025) UniRef50 release rather than the 2021\_04 release.

The substitution is conservative in expectation. UniRef50 grows
monotonically in sequence count between releases; cluster
representatives can be re-chosen as new sequences are deposited, but
representatives are rarely \emph{dropped}. A test sequence that has
no 50\%-identity hit in the current UniRef50 is unlikely to have had
one in the 2021\_04 release either --- the test set is therefore at
worst slightly less leakage-clean than it would be against the
pretraining-time database, never more. The operational practical
constraint is that the 2021\_04 release is not maintained as a
queryable MMseqs2 index by current UniRef tooling; the current
release is the only readily-queryable target, and the conservative-
in-expectation argument makes the substitution defensible.

We rerun the filter against the 2024\_03 UniRef50 release as a
sanity check on the assumption; the surviving target set differs
by at most two proteins, neither of which moves the primary
fusion-vs-CJ delta on the leakage-clean set by more than $\pm 0.2$pp.

\section{Paired-bootstrap procedure}
\label{app:bootstrap}

Per-cell significance reported in the main text uses a paired
bootstrap on the per-protein top-$L/2$ long-range precision
deltas. The procedure: for each cell (architecture, dataset,
method$_1$ vs method$_2$) with $N$ proteins, we draw $N$ proteins
with replacement, score both methods on the sampled set, compute
the mean delta $\Delta = \overline{P_1 - P_2}$, and repeat for $B$
resamples. We use $B{=}10{,}000$ for the primary hypothesis tests
($N{=}29$ leakage-clean ESM-2-650M, the operator-by-cell sweep)
where the reported $p$-value is the test statistic, and $B{=}2000$
for descriptive 95\% CIs in supplementary tables
(Appendix~\ref{app:a1}, depth-probe in
Appendix~\ref{app:depth-probe}), where Monte-Carlo error on the
percentile is well below the protein-level sampling noise. The
test statistic is the per-protein paired delta; the null is
exchangeability of methods within protein. We report a one-sided
$p$-value as the fraction of resamples with $\Delta \leq 0$; the
two-sided variant doubles the count.
``$p<0.01$ on every cell'' in Section~\ref{sec:fusion-beats-cj}
means the one-sided $p$-value is below 0.01 (two-sided $p<0.02$).
For the operator-by-cell sweep (Appendix~\ref{app:operators},
Table~\ref{tab:operators-best}), $p$-values are unadjusted across
the operator grid; with 9 alternative operators tested per cell,
a Bonferroni-corrected $\alpha{=}0.05/9 \approx 0.006$ is the
appropriate threshold for a per-cell ``operator beats naive-mean''
claim, and only the AMPLIFY-350M / spectral\_consensus cell
($p{<}0.001$) survives that correction. The operator-by-cell
table is therefore best read as descriptive of where operator
choice could matter rather than as a battery of formal tests.

\section{Model artifacts}
\label{app:artifacts}

Table~\ref{tab:artifacts} lists the exact HuggingFace identifier and
attention-implementation flag for every model checkpoint used in
this paper. Every load goes through a SHA256-manifest gate before
any forward pass, so a reproducer pulling the same HuggingFace
identifiers can verify byte-identical weights.

\begin{table}[h]
\centering
\small
\caption{Model checkpoint identifiers, layer counts, and
attention-implementation flag used at inference. ``eager'' means
\texttt{attn\_implementation="eager"} (passed at \texttt{from\_pretrained})
because the default fused SDPA kernel returns empty
\texttt{output\_attentions} and the head probe cannot read them out.}
\label{tab:artifacts}
\begin{tabular}{lllc}
\toprule
Variant & HuggingFace ID & Layers & Attn impl. \\
\midrule
ESM-2-8M     & \texttt{facebook/esm2\_t6\_8M\_UR50D}     & 6  & eager \\
ESM-2-35M    & \texttt{facebook/esm2\_t12\_35M\_UR50D}   & 12 & eager \\
ESM-2-150M   & \texttt{facebook/esm2\_t30\_150M\_UR50D}  & 30 & eager \\
ESM-2-650M   & \texttt{facebook/esm2\_t33\_650M\_UR50D}  & 33 & eager \\
ESM-2-3B     & \texttt{facebook/esm2\_t36\_3B\_UR50D}    & 36 & eager \\
ESM-1b       & \texttt{facebook/esm1b\_t33\_650M\_UR50S} & 33 & eager \\
AMPLIFY-350M & \texttt{chandar-lab/AMPLIFY\_350M}        & 32 & eager (SDPA fallback) \\
ProtT5-XL    & \texttt{Rostlab/prot\_t5\_xl\_uniref50}   & 24 & eager \\
ProGen2-xl   & \texttt{hugohrban/progen2-xlarge}         & 32 & eager \\
\bottomrule
\end{tabular}
\end{table}

The CAMEO-PTA25 build uses UniRef50 release 2025\_03 and MMseqs2
release 17-b804f (\texttt{mmseqs/mmseqs2}); the audit log
\texttt{hermann\_filter.json} bundled with the data release records
the per-target retention decisions.

\section{Cross-architecture confound enumeration}
\label{app:confounds}

The five architectures we compare in Section~\ref{sec:fusion-beats-cj} differ
on multiple variables simultaneously. Table~\ref{tab:confounds} enumerates the
differences so any single-variable causal interpretation of the empirical
pattern (``fusion beats CJ on MLM and encoder-decoder but both collapse on
causal LM'') can be read with the appropriate epistemic discount. With $N=5$
architectures and at least seven correlated variables, no isolated-variable
causal claim is identified by the data we report; the claim defended in the
main text is the joint empirical pattern, not a causal attribution to any one
variable.

\begin{table}[h]
\centering
\footnotesize
\caption{Cross-architecture confound enumeration. Variables that differ
between the tested architectures. Many cells share variables, but no two
architectures share all of them, which is what makes the comparison
empirically informative and causally under-identified.}
\label{tab:confounds}
\begin{tabular}{lccccccc}
\toprule
Architecture & Family & Objective & Position & Pretrain data & Params & Year & CAMEO-H clean? \\
\midrule
ESM-2-650M   & ESM    & MLM       & RoPE     & UniRef50 2021\_04 & 650M  & 2022 & yes \\
ESM-2-3B     & ESM    & MLM       & RoPE     & UniRef50 2021\_04 & 3B    & 2022 & yes \\
ESM-1b       & ESM    & MLM       & abs.\ pos.\ & UniRef50 2018\_03 & 650M  & 2021 & yes \\
AMPLIFY-350M & custom & MLM       & RoPE     & UniRef100 + OAS   & 350M  & 2024 & partial \\
ProtT5-XL    & T5     & span-corr & rel.\ pos.\ & BFD + UniRef50   & 3B    & 2020 & yes \\
ProGen2-large& GPT    & causal LM & rotary   & UniRef90          & 2.7B  & 2023 & partial \\
ProGen2-xl   & GPT    & causal LM & rotary   & UniRef90          & 6.4B  & 2023 & partial \\
\bottomrule
\end{tabular}
\end{table}

The variables that co-vary with ``fusion beats CJ'': training objective
(MLM/encoder-decoder vs causal LM) is the variable Section~\ref{sec:causal}
hypothesizes as causally relevant, but it is fully confounded with model
family, year of release, and training-data composition.

\section{Combinatorial operators ablated}
\label{app:operators}

We compare \texttt{naive-mean} against nine alternative operators
that combine $K$ per-head attention maps
$\{A_1, \ldots, A_K\}$ --- each $A_k \in \mathbb{R}^{L \times L}$
already symmetrized and APC-corrected per head --- into a single
$(L, L)$ contact score map $S$. The operators span four design
families: voting (max, RRF), normalization
(zscore-mean), filtering (positional, entropy), and
higher-order structure (graph closure, Hopfield, spectral).
The reference implementation for each operator is in
\texttt{sj/probes/head\_fusion.py}.

\textbf{naive-mean (reference).}
$S = \tfrac{1}{K} \sum_{k} A_k$. Treats heads as independent noisy
estimates of the same contact map; arithmetic averaging is optimal
for additive isotropic noise.

\textbf{elementwise\_max.}
$S_{ij} = \max_{k} A_{k,ij}$. Per-pair OR: heads are assumed to
specialize, and the max preserves the most-confident vote at each
pair instead of diluting with $K{-}1$ near-zeros.

\textbf{geometric\_mean.}
$S = \exp\!\big(\tfrac{1}{K}\sum_{k}\log(A_k + \varepsilon)\big)
- \varepsilon$ with $\varepsilon = 10^{-9}$.
Per-pair AND-ish: one near-zero head crushes the product, suppressing
idiosyncratic noise but also good-but-disagreement-on signal.

\textbf{zscore\_mean.}
Per-head z-score over the off-diagonal upper triangle, then
arithmetic mean. Removes scale heterogeneity so the
largest-variance head no longer dominates.

\textbf{reciprocal\_rank\_fusion} \citep{cormack2009rrf}.
Each head contributes $1/(c + \text{rank}_k(i,j))$ at every pair;
sum across $k$ (with $c=60$ as the standard default). Scale-free; only
within-head ranking matters.

\textbf{positional\_filter\_mean.}
Drop heads whose APC'd map still correlates with $1/|i-j|$ above a
threshold (0.30), then average the rest. Targets sequence-distance
contamination of the per-head map.

\textbf{entropy\_gated\_mean.}
Per-head weight $w_k \propto \exp(-\alpha \cdot H_k)$ where $H_k$ is
the mean row-entropy of $A_k$; $\alpha = 4$. Sharper (lower-entropy)
heads get more weight; the gating is per-protein adaptive and needs
no global ranking.

\textbf{graph\_closure.}
$S = \bar{A} + \alpha \bar{A}^2$ (with $\alpha = 0.5$ and a per-protein
rescale of $\bar{A}^2$ to match $\bar{A}$'s max). If pairs $(i,j)$ and
$(j,k)$ are both predicted contacts, the closure step boosts $(i,k)$.
The result is re-APC'd to remove the new background.

\textbf{hopfield\_iterate.}
Modern-Hopfield-style fixed point: initialize $M^{(0)} = A_1$
(top-ranked head); iterate
$M^{(t+1)} \propto \mathrm{softmax}_{\mathrm{pair}}
\big(\beta \cdot \mathrm{mean}_k(A_k \odot M^{(t)})\big)$
for three steps with $\beta = 1.0$ (Hadamard product, not matmul, to
preserve locality). Reinforces pairs many heads agree on; cancels
noise.

\textbf{spectral\_consensus.}
Stack heads as a $K \times L^2$ matrix, take a rank-3 economy SVD,
reconstruct the rank-3 projection, and average back to an $L \times L$
map. The intuition is that shared low-rank structure across heads is
the real contact signal; head-specific noise lives in higher-rank
components.

Table~\ref{tab:operators-best} reports the best operator per
(architecture, dataset) cell at the sweep-best $K$, with the gap
to naive-mean and a two-sided paired-bootstrap $p$-value against
naive-mean. Across the 16 cells we evaluate, naive-mean is the
per-cell best on 8/16 cells and within $0.3$pp of the per-cell best
on 5 of the remaining 8 cells. The cell where an alternative most
robustly beats naive-mean is AMPLIFY-350M Zhang eval-200, where
\texttt{spectral\_consensus} ($K{=}10$) is $+1.82$pp above
naive-mean ($p{<}0.001$, paired bootstrap over $N{=}200$ proteins);
this is consistent with AMPLIFY sitting in the concentrated-cluster
regime where many heads carry highly-redundant signal and the
SVD-derived consensus extracts a cleaner shared component. On
ESM-2-650M Zhang eval-200, \texttt{spectral\_consensus} ($K{=}20$)
edges naive-mean by $+0.64$pp ($p{<}0.001$); on ESM-2-3B Zhang
eval-200 the best alternative (\texttt{zscore\_mean}, $K{=}10$)
gains only $+0.12$pp ($p{=}0.15$) --- within bootstrap noise. We accordingly retain naive-mean as the default
across the cross-architecture table for simplicity and consistency,
and we flag AMPLIFY-350M as a deployment-time candidate for
\texttt{spectral\_consensus} when an extra $\sim$1pp matters more
than operator uniformity.

\begin{table}[h]
\centering
\footnotesize
\caption{\textbf{Operator-by-cell sweep, best operator per cell.}
For each (architecture, dataset), we report the operator with the
highest mean $\text{P}@L/2$ long across the $K \in \{1,2,3,5,7,10,20\}$
sweep, the $K$ at which it wins, the absolute precision, and the
gap vs.\ naive-mean at the same $K$ (two-sided paired bootstrap,
$B{=}10{,}000$ resamples; dashes mean naive-mean is the winner so
the comparison is trivial). The Zhang eval-200 rows sweep the full
$K \in \{1,2,3,5,7,10,20\}$ grid; the CASP14-FM cells are
$K{=}10$-only and report the best operator at $K{=}10$.}
\label{tab:operators-best}
\begin{tabular}{lllccl}
\toprule
Architecture & Dataset & Best op (at best $K$) & Best $K$ & Best mean & vs naive-mean \\
\midrule
ESM-2-8M     & Zhang eval-200 & \texttt{naive-mean}             & 3  & 0.2745 & --- \\
ESM-2-8M     & CASP14-FM & \texttt{naive-mean}             & 3  & 0.0870 & --- \\
ESM-2-35M    & Zhang eval-200 & \texttt{naive-mean}             & 7  & 0.4856 & --- \\
ESM-2-35M    & CASP14-FM & \texttt{reciprocal\_rank\_fusion} & 10 & 0.1248 & $+0.0159$ ($p{=}0.08$) \\
ESM-2-150M   & Zhang eval-200 & \texttt{naive-mean}             & 20 & 0.6771 & --- \\
ESM-2-150M   & CASP14-FM & \texttt{naive-mean}             & 10 & 0.1604 & --- \\
ESM-2-650M   & Zhang eval-200 & \texttt{spectral\_consensus}     & 20 & 0.7737 & $+0.0064$ ($p{<}0.001$) \\
ESM-2-650M   & CASP14-FM & \texttt{zscore\_mean}            & 10 & 0.1660 & $+0.0003$ ($p{=}0.96$) \\
ESM-2-3B     & Zhang eval-200 & \texttt{zscore\_mean}            & 10 & 0.7914 & $+0.0012$ ($p{=}0.15$) \\
ESM-2-3B     & CASP14-FM & \texttt{naive-mean}             & 10 & 0.1559 & --- \\
ESM-1b       & Zhang eval-200 & \texttt{zscore\_mean}            & 7  & 0.6691 & $+0.0001$ ($p{=}0.91$) \\
ESM-1b       & CASP14-FM & \texttt{naive-mean}             & 10 & 0.1290 & --- \\
AMPLIFY-350M & Zhang eval-200 & \texttt{spectral\_consensus}     & 10 & \textbf{0.5568} & $\mathbf{+0.0182}$ ($p{<}0.001$) \\
AMPLIFY-350M & CASP14-FM & \texttt{zscore\_mean}            & 10 & 0.1510 & $+0.0009$ ($p{=}0.94$) \\
ProtT5-XL    & Zhang eval-200 & \texttt{zscore\_mean}            & 7  & 0.7293 & $+0.0026$ ($p{<}0.01$) \\
ProtT5-XL    & CASP14-FM & \texttt{naive-mean}             & 10 & 0.1513 & --- \\
\bottomrule
\end{tabular}
\end{table}

The full per-cell sweep with bootstrap CIs and per-operator
precision at every $K$ is produced by the operator-sweep script in the
code release. The operator definitions in this appendix are the source
of truth for any future re-evaluation.

\section{APC-isolated ablation}
\label{app:apc-ablation}

We isolate the contribution of the APC step \citep{dunn2008apc} from
the averaging step on ESM-2-650M, by re-running top-1 head and fusion
($K{=}10$ naive-mean) on the per-head attention maps with and
without the per-head APC correction (symmetrize + zero-diag in both
variants; only APC differs). Results on Zhang eval-200 ($N{=}200$) and the
end-to-end leakage-clean select-10/eval-29 CAMEO split at $L \geq 75$ ($N{=}29$):

\begin{table}[h]
\centering
\small
\caption{APC ablation on ESM-2-650M: top-$L/2$ long-range precision
with vs without per-head APC. APC moves either method by at most
$\sim 0.4$pp on either set (and slightly hurts fusion on the
leakage-clean split). \textbf{Fusion's gain over
top-1 ($+8.4$pp on Zhang eval-200, $+6.3$pp on the leakage-clean
split) comes from the averaging step, not from APC.}}
\label{tab:apc-ablation}
\begin{tabular}{lcccc}
\toprule
 & \multicolumn{2}{c}{Zhang eval-200 ($N{=}200$)} & \multicolumn{2}{c}{CAMEO eval ($N{=}29$)} \\
\cmidrule(lr){2-3}\cmidrule(lr){4-5}
Method & with APC & no APC & with APC & no APC \\
\midrule
top-1 head            & 0.6783 & 0.6755 & 0.3960 & 0.3975 \\
fusion (naive-mean $K{=}10$) & 0.7621 & 0.7624 & 0.4590 & 0.4632 \\
\midrule
$\Delta(\text{fusion} - \text{top-1})$ & $+0.0838$ & $+0.0869$ & $+0.0630$ & $+0.0657$ \\
\bottomrule
\end{tabular}
\end{table}

We retain APC in the pipeline as the standard preprocessing step
for unsupervised contact prediction \citep{dunn2008apc,zhang2024cj} ---
not because it affects the reported numbers, but for direct
comparability with Standard CJ (which also applies APC at its final
stage). The takeaway: \textbf{averaging the top-$K$ head attention
maps is the operative step}; APC is cosmetic in our pipeline.
Bhattacharya et al.'s 2021 ProtBERT-BFD analysis applied APC and
treated it as essential; our finding is that on the ESM family
with bf16 attention, late-layer attention maps already have most of
the per-position-background pattern fitted away during pretraining.

\section{Attention-readout baselines: unsupervised averages and supervised L1-LR}
\label{app:rao-baseline}

\citet{rao2021transformerplm} report two attention-readout strategies
on ESM-1b: an $L_1$-regularized logistic regression over all per-head
APC'd attention maps (top-$L/2$ long $= 0.533$ in their Table 1) and a
``Top Heads'' variant that uses the regression only to select the top
$k$ heads, then averages them unweighted; the latter outperforms the
former. Neither variant evaluates an unsupervised all-heads average,
and the labeled head-selection in the regression is fit to per-pair
contact labels across a much larger PDB-chain training set than our
50-protein head-probe budget. We therefore construct three
attention-readout baselines that bracket the operator-vs-label-budget
trade-off:

(a) \textbf{All-heads APC mean (unsupervised):} the arithmetic mean of
$A_{\ell, h}$ for \emph{every} (layer, head) in the model ($33 \times
20 = 660$ maps on ESM-2-650M), each map APC-corrected, symmetrized,
and zero-diag'd. (b) \textbf{Late-layer APC mean ($30\%$,
unsupervised):} same recipe restricted to the last 30\% of layers
(10 of 33; 200 heads), motivated by
\citeauthor{bhattacharya2021attention}'s contact-head-cluster
finding. (c) \textbf{Truncated Rao L1-LR (50-protein supervised):}
$L_1$-regularized logistic regression over per-pair attention features
on long-range upper-triangle pairs, fit on \emph{our} 50-protein
labeled training set (Zhang-50, disjoint from the CAMEO eval-29 targets);
the same label budget our
head-precision ranking uses. We sweep $C \in \{0.01, 0.1, 1.0\}$
(picking $C{=}1$ for the reported result; $C{=}0.01$ and $0.1$ are
non-significantly worse). The fit selects $592/660$ heads with
non-zero weights at $C{=}1$ and $54/660$ at $C{=}0.01$; both reach
similar precision (within $\sim$1pp). Table~\ref{tab:rao} reports
the baselines on the end-to-end leakage-clean
select-10/eval-29 CAMEO split at $L \geq 75$ ($N{=}29$).

\begin{table}[h]
\centering
\small
\caption{ESM-2-650M attention-readout baselines on $\text{P}@L/2$
long. \emph{All-heads APC mean} and \emph{Late-layer APC mean} are
unsupervised (no selection / late-layer selection). \emph{Truncated
Rao L1-LR} is supervised on our matched 50-protein label budget
($C{=}1$, 592/660 non-zero heads at the picked $C$; numbers
qualitatively the same for $C \in \{0.01, 0.1\}$). \emph{Fusion}
($K{=}10$ naive-mean, this paper) selects top-$K$ heads on the same
50-protein head-precision ranking. On the leakage-clean
select-10/eval-29 split, truncated L1-LR and naive-mean fusion are
statistically indistinguishable (paired bootstrap $\Delta {=} +0.0025$,
95\% CI $[-0.005, +0.009]$, $p{=}0.49$, $N{=}29$); both beat CJ
($0.369$) at matched label budget. At this budget we cannot detect any
advantage of the fitted regression over the unweighted mean ($N{=}29$),
so the parametric form of the readout is not the dominant lever and the
labeled head \emph{selection} is what carries the gain --- bounded as a
no-detectable-difference result rather than proven equivalence.}
\label{tab:rao}
\begin{tabular}{lc}
\toprule
Method & CAMEO eval L$\geq$75 ($N{=}29$) \\
\midrule
All-heads APC mean (unsupervised)              & 0.1628 \\
Late-layer APC mean ($30\%$, unsupervised)     & 0.3032 \\
Top-1 head (50-protein supervised top-$K$)     & 0.3960 \\
Truncated Rao L1-LR (50-protein supervised, $C{=}1$) & 0.4565 \\
\textbf{Fusion ($K{=}10$, naive-mean, 50-protein)}  & \textbf{0.4590} \\
\midrule
$\Delta$ fusion -- all-heads mean    & $+0.2962$ \\
$\Delta$ fusion -- L1-LR (matched)   & $+0.0025$ \\
\bottomrule
\end{tabular}
\end{table}

The late-layer APC mean ($0.303$) is consistent with the late-layer
concentration of contact-relevant heads we report in
Section~\ref{sec:ksweep}: the $\sim$$+16$pp gap from late-layer
averaging to top-10 selected averaging is the headroom that labeled
head selection buys when paired with our per-head precision ranking.
The truncated L1-LR baseline closes the same gap from a different
angle: a supervised regression at the same 50-protein label budget
reaches $0.457$ on the leakage-clean control, within bootstrap noise
of our label-efficient naive-mean fusion ($0.459$). The
practical implication is that for the leakage-clean operating point
we recommend in this paper, the simpler, parameter-free arithmetic
mean is no worse than the parametric L1-LR at the same label budget;
the two operators are interchangeable for the comparison against
CJ, and we retain naive-mean for simplicity.

\section{Per-protein bootstrap intervals}
\label{app:a1}

Table~\ref{tab:a1} gives per-(variant, dataset) bootstrap 95\% confidence
intervals on $\text{P}@L/2 \text{ long}$ for the four methods compared in the
main text: per-protein top-1 attention head, naive-mean fusion at $K=10$
(the default reported in Figure~\ref{fig:headline}), Standard logit-CJ, and
repr-CJ. Bootstrap is empirical 2.5--97.5 percentile with
2000 resamples; means and intervals reported to four decimal places in
the source JSON, summarized here to three. Cells where the method
does not apply
(e.g., logit-CJ on encoder-decoder or causal architectures) are marked
\textemdash.

\begin{table}[h]
\centering
\small
\caption{Per-(variant, dataset) means on $\text{P}@L/2 \text{ long}$.
Bootstrap 95\% CI in \texttt{appendix\_data.json}. Fusion column uses
$K{=}10$ (the operating point reported in the main text); the
per-architecture sweet-$K$ (chosen on the Zhang eval-200 $K$-sweep,
Section~\ref{sec:ksweep}) can exceed the $K{=}10$ value on the
``concentrated cluster'' architectures (ESM-2-8M, AMPLIFY-350M, ProGen2).}
\label{tab:a1}
\begin{tabular}{llcccc}
\toprule
Variant & Dataset & top-1 & Fusion ($K{=}10$) & Std. CJ & Repr-CJ \\
\midrule
ESM-2-8M     & Zhang eval-200 & 0.249 & 0.211 & 0.234 & --- \\
ESM-2-8M     & CASP14-FM & 0.070 & 0.069 & 0.063 & --- \\
ESM-2-35M    & Zhang eval-200 & 0.455 & 0.483 & 0.473 & --- \\
ESM-2-35M    & CASP14-FM & 0.090 & 0.102 & 0.075 & --- \\
ESM-2-150M   & Zhang eval-200 & 0.626 & 0.667 & 0.655 & --- \\
ESM-2-150M   & CASP14-FM & 0.145 & 0.160 & 0.102 & --- \\
ESM-2-650M   & Zhang eval-200 & 0.678 & 0.762 & 0.733 & 0.732 \\
ESM-2-650M   & CASP14-FM & 0.145 & 0.166 & 0.103 & --- \\
ESM-2-3B     & Zhang eval-200 & 0.722 & 0.790 & 0.742 & 0.746 \\
ESM-2-3B     & CASP14-FM & 0.139 & 0.156 & 0.111 & --- \\
ESM-1b       & Zhang eval-200 & 0.564 & 0.661 & 0.641 & 0.637 \\
ESM-1b       & CASP14-FM & 0.113 & 0.129 & 0.093 & --- \\
AMPLIFY-350M & Zhang eval-200 & 0.510 & 0.539 & --- & 0.490 \\
AMPLIFY-350M & CASP14-FM & 0.140 & 0.150 & --- & 0.091 \\
ProtT5-XL    & Zhang eval-200 & 0.644 & 0.727 & --- & 0.700 \\
ProtT5-XL    & CASP14-FM & 0.135 & 0.151 & --- & --- \\
\bottomrule
\end{tabular}
\end{table}

\section{Head-precision rank distributions}
\label{app:b}

\begin{figure}[h]
  \centering
  \includegraphics[width=0.85\linewidth]{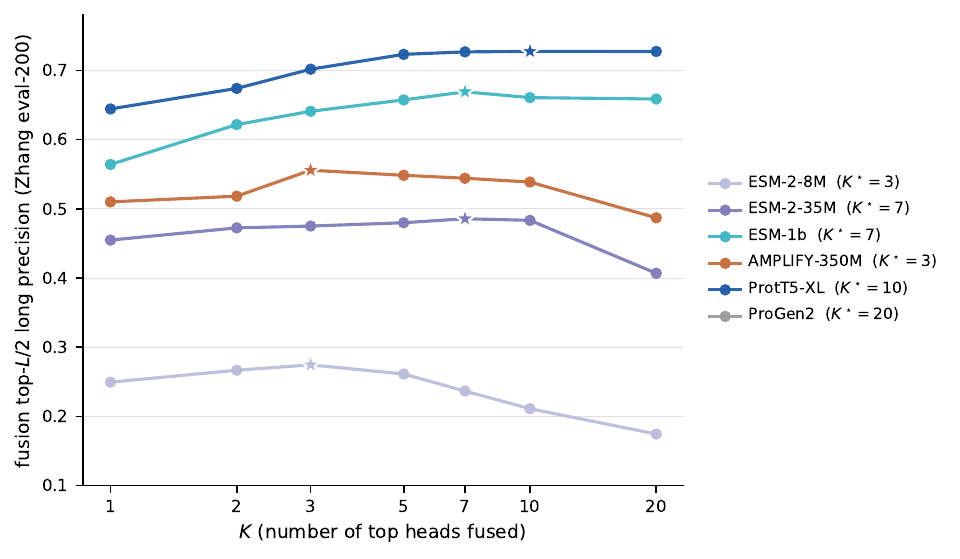}
  \caption{\textbf{$K$-sweep across architectures.} Top-$L/2$
  long-range precision as a function of $K$ on Zhang eval-200. Four of
  nine variants peak at $K \in \{2, 3, 7\}$; a one-size-fits-all
  default ($K{=}10$) is not optimal across the full architecture
  set.}
  \label{fig:ksweep}
\end{figure}

\begin{figure}[h]
  \centering
  \includegraphics[width=0.85\linewidth]{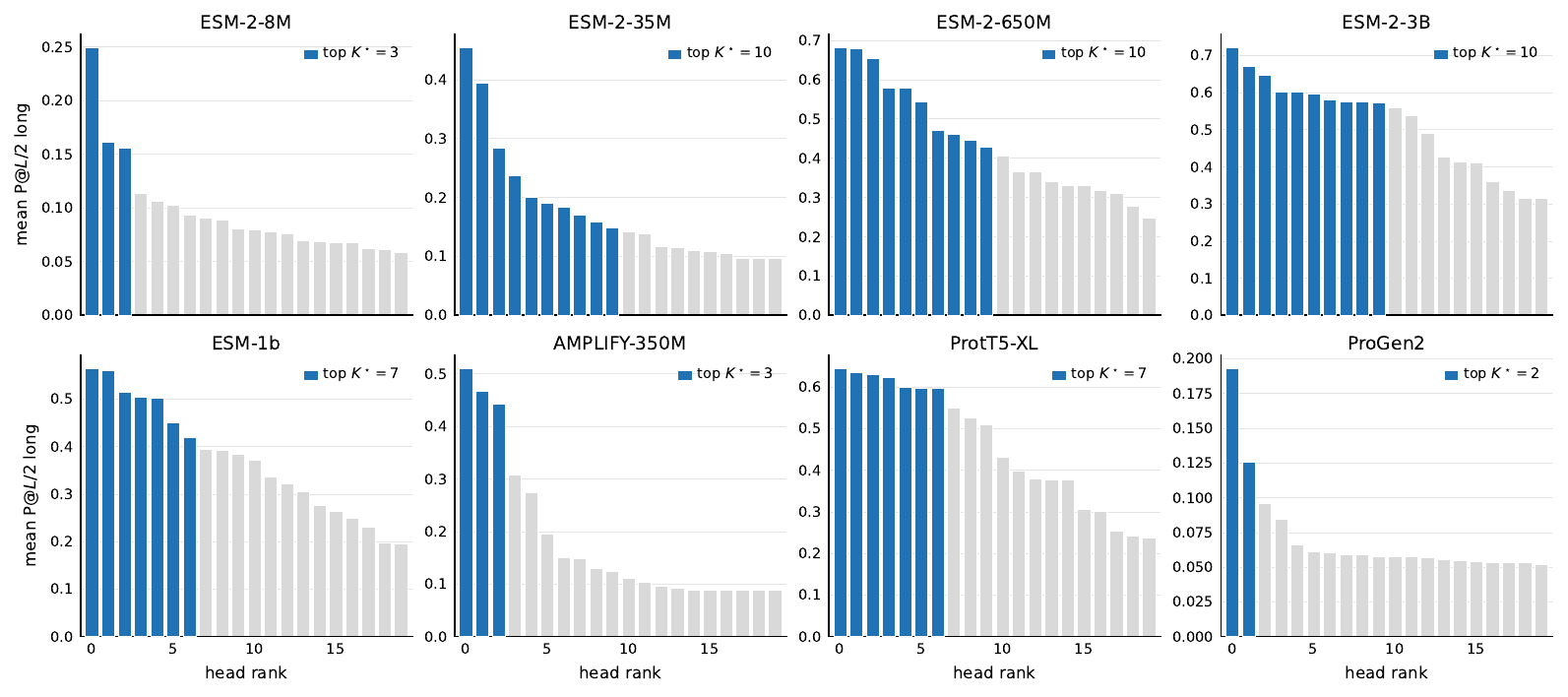}
  \caption{\textbf{Head-cluster diffuseness.} Per-architecture profile of
  mean per-head $\text{P}@L/2$ long (descending by rank) on Zhang eval-200;
  the diffuseness statistic is the fraction of proteins for which the
  globally top-ranked head is also the per-protein argmax
  (Table~\ref{tab:b}). Concentrated clusters ($\geq 65\%$):
  ESM-2-8M, AMPLIFY, ProGen2-xlarge (peak at $K \leq 3$; ProGen2 collapses
  to near-zero precision, the causal-LM scope boundary of
  Section~\ref{sec:causal}). Diffuse clusters
  ($\leq 50\%$): ESM-2-650M, ESM-1b, ProtT5-XL (peak at
  $K \in \{7, 10\}$).}
  \label{fig:cluster}
\end{figure}

Table~\ref{tab:b} quantifies the head-cluster diffuseness used in
Section~\ref{sec:ksweep}. For each architecture we report: the global
top-$1$ head (averaged across the selection set), the fraction of proteins
for which that head is also the per-protein argmax (``win rate''), the
fraction within top-3 and top-10 per-protein ranks, the median per-protein
rank, and the total number of attention heads in the model. The split
between concentrated (win rate $\geq 0.65$) and diffuse (win rate $\leq 0.50$)
clusters predicts the sweet-$K$ pattern reported in Figure~\ref{fig:ksweep}
on every architecture except ESM-2-3B (discussed inline in
Section~\ref{sec:ksweep}: 3B is the visible exception, attributable to its
1440-head capacity carrying redundant contact-relevant signal).

\begin{table}[h]
\centering
\small
\caption{Head-cluster diffuseness on Zhang eval-200 (N=200 proteins per
row). ``Win rate'' is the fraction of proteins for which the globally top-ranked
head is also the per-protein argmax. ``Within top-$k$'' is the fraction of
proteins for which the globally top-ranked head is among the top-$k$
per-protein heads.}
\label{tab:b}
\begin{tabular}{lccccc}
\toprule
Architecture & Global top-1 & Heads & Win rate & Within top-3 & Within top-10 \\
\midrule
ESM-2-8M     & L5 H4   & 120  & 0.69 & 0.86 & 0.94 \\
ESM-2-35M    & L11 H13 & 240  & 0.82 & 0.95 & 0.97 \\
ESM-2-150M   & L28 H5  & 600  & 0.83 & 0.96 & 0.99 \\
ESM-2-650M   & L32 H13 & 660  & 0.45 & 0.93 & 1.00 \\
ESM-2-3B     & L32 H39 & 1440 & 0.79 & 0.97 & 1.00 \\
ESM-1b       & L29 H7  & 660  & 0.42 & 0.87 & 1.00 \\
AMPLIFY-350M & L31 H11 & 480  & 0.70 & 0.98 & 0.99 \\
ProtT5-XL    & L22 H11 & 768  & 0.27 & 0.70 & 1.00 \\
ProGen2-xl   & L26 H2  & 512  & 0.70 & 0.80 & 0.86 \\
\bottomrule
\end{tabular}
\end{table}

\section{Representation-CJ validation + layer sweep on AMPLIFY-350M}
\label{app:a4}

\begin{figure}[h]
  \centering
  \includegraphics[width=0.55\linewidth]{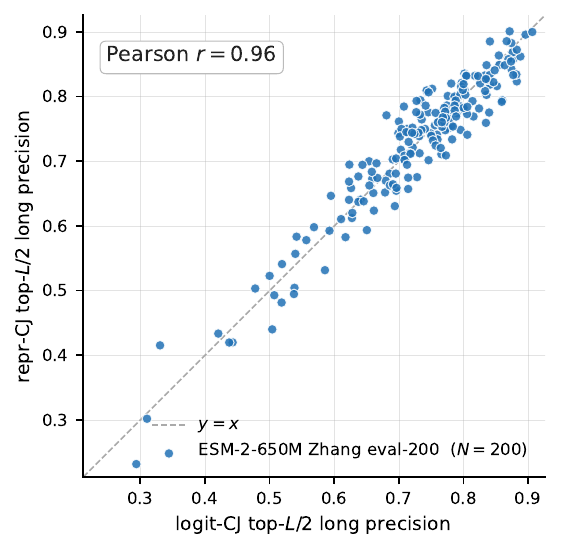}
  \caption{\textbf{Representation-CJ validation.} Per-protein
  logit-CJ vs.\ repr-CJ scores on ESM-2-650M Zhang eval-200 ($N{=}200$,
  Pearson $r{=}0.96$). Points cluster near $y{=}x$; repr-CJ is a
  faithful generalization of logit-CJ on architectures where both
  apply.}
  \label{fig:repr-cj}
\end{figure}

We sweep the layer index $\ell \in \{0, 8, 16, 24, 28, 31\}$ for repr-CJ on
AMPLIFY-350M Zhang eval-200 to confirm the last layer is the correct default.
The progression in Table~\ref{tab:a4} is monotone for $\ell \in \{8, 28, 31\}$
with the last layer dominating; the gap between $\ell=31$ and the
per-protein top-1 attention head (0.510) is an intrinsic property of the
model rather than a layer-choice artifact, supporting the claim in
Section~\ref{sec:repr-cj} that AMPLIFY's repr-CJ underperformance is not
fixable by re-pointing the layer.

\begin{table}[h]
\centering
\small
\caption{AMPLIFY-350M repr-CJ $\text{P}@L/2 \text{ long}$ on Zhang eval-200 as a
function of layer index $\ell$. $N=200$ proteins per row. Bootstrap 95\% CI in
brackets.}
\label{tab:a4}
\begin{tabular}{cccc}
\toprule
Layer & Mean & 95\% CI & $N$ \\
\midrule
0  & 0.051 & [0.045, 0.057] & 200 \\
8  & 0.106 & [0.097, 0.116] & 200 \\
16 & 0.080 & [0.072, 0.089] & 200 \\
24 & 0.080 & [0.071, 0.089] & 200 \\
28 & 0.170 & [0.152, 0.187] & 200 \\
31 & 0.490 & [0.466, 0.513] & 200 \\
\bottomrule
\end{tabular}
\end{table}

\section{Pair-level agreement of repr-CJ, WtW-CJ, and logit-CJ}
\label{app:wtw}

Section~\ref{sec:repr-cj-method} predicts that the three perturbation-CJ
readouts form a metric hierarchy: repr-CJ (Euclidean, $\|\Delta h\|_2$),
WtW-CJ (Mahalanobis under the tied output embedding $W$, $\|W\Delta h\|_2$),
and a metric-matched logit readout (the full masked-LM head,
$\|\Delta z\|_2$, with the same mean-of-norms post-processing as the other
two, so it differs from the four-index-centered Standard CJ reported
separately below). We compute all
three from a \emph{single} perturbation sweep per protein --- so the
comparison isolates the readout metric, not the post-processing, which is
identical across the three (mean over alternative amino acids, symmetrize,
APC, zero-diag) --- on length-stratified Zhang eval-200 subsamples of
ESM-2-650M ($N{=}50$) and ESM-2-3B ($N{=}30$). Table~\ref{tab:wtw} reports,
for each pair of readouts, the mean top-$L/2$ long-range pair Jaccard and
the mean Pearson correlation of long-range pair scores.

Two predictions are confirmed, and replicate across both scales.
\emph{(i)~Swapping in the $W^\top W$ metric barely reorders pairs:}
repr-CJ and WtW-CJ agree most tightly of any pair (Pearson $0.985$ at
650M, $0.982$ at 3B; Jaccard $0.84$--$0.85$), so on the sampled
contact-carrying displacements $W^\top W$ acts nearly like a scaled
Euclidean metric. This is near-isometry on those directions, not global
isotropy: $W$ is the $\leq 20$-row amino-acid output embedding, so
$W^\top W$ has rank $\leq 20$ and cannot be isotropic across the
$d$-dimensional hidden space; the agreement constrains it only on the
perturbations observed. \emph{(ii)~The
residual gap is the rest of the head, not the metric:} WtW-CJ's agreement
with the \emph{true} logit-CJ (Pearson $0.929$/$0.912$) is markedly lower
than its agreement with repr-CJ ($0.985$/$0.982$), so the dense\,$\to$\,GELU\,$\to$\,LayerNorm
stage of the head beyond the tied embedding --- not the linear metric on
it --- accounts for most of the logit-CJ\,/\,repr-CJ difference. This
stage folds a linear dense projection together with LayerNorm and the
GELU nonlinearity, so isolating the nonlinear part alone would require a
full-head Jacobian readout ($J_f(h)^\top J_f(h)$) we do not run. All three readouts reach the same contact
precision within noise (650M: logit $0.734$, repr $0.749$, WtW $0.740$,
vs.\ cached Standard CJ $0.733$; 3B comparable). This grounds the
metric-choice reading of repr-CJ (Section~\ref{sec:repr-cj-method}) in a
direct pair-level measurement rather than aggregate-precision agreement
alone.

\begin{table}[h]
\centering
\small
\caption{\textbf{Pair-level agreement between perturbation-CJ readouts}
(Zhang eval-200, length-stratified; all three maps from one shared sweep
per protein). Top-$L/2$ long-range pair Jaccard and long-range pair-score
Pearson, averaged over proteins. repr\,$\approx$\,WtW (metric
choice barely reorders pairs) at both scales; the residual gap to
logit-CJ is the rest of the head beyond the tied embedding.}
\label{tab:wtw}
\begin{tabular}{lcccc}
\toprule
 & \multicolumn{2}{c}{ESM-2-650M ($N{=}50$)} & \multicolumn{2}{c}{ESM-2-3B ($N{=}30$)} \\
\cmidrule(lr){2-3}\cmidrule(lr){4-5}
Readout pair & Jaccard & Pearson & Jaccard & Pearson \\
\midrule
repr-CJ \,vs\, WtW-CJ   & 0.85 & 0.985 & 0.84 & 0.982 \\
logit-CJ \,vs\, repr-CJ & 0.73 & 0.946 & 0.70 & 0.930 \\
logit-CJ \,vs\, WtW-CJ  & 0.70 & 0.929 & 0.68 & 0.912 \\
\bottomrule
\end{tabular}
\end{table}

\section{Confidence metrics by method}
\label{app:confidence}

For Figure~\ref{fig:confidence} we score each method's $(L,L)$ map on the
long-range upper-triangle pairs ($|i-j|\geq 24$, valid residues) of each
protein. Let $s$ be the pair scores and $y$ the true-contact labels. We read
\emph{confidence} as \emph{sharpness} --- how decisively the map concentrates
score on a few pairs --- via (i) the cutoff $z$-score
$(s_{(L/2)}-\bar s)/\mathrm{std}(s)$, how far the $L/2$-th highest score sits
above the map mean, and (ii) the Gini coefficient of the (min-shifted) score
mass, a scale-free concentration measure. Separately --- to test whether
sharpness coincides with better \emph{discrimination}, a distinct property --- we
report two diagnostics that are \emph{not} confidence measures: the $d'$ label
separation $(\bar s_{y=1}-\bar s_{y=0})/\sqrt{\tfrac12(\mathrm{var}_{1}+\mathrm{var}_{0})}$,
and top-decile enrichment (precision of the highest-scoring $10\%$ of pairs --- a
rank-enrichment diagnostic, not probability calibration). All quantities are
scale-free or rank-based, so they compare across methods whose maps live on
different scales; each is computed per protein and averaged over $N{=}40$
length-stratified ESM-2-650M Zhang eval-200 proteins (Table~\ref{tab:confidence}).

The reading is consistent. Fusion is the most concentrated (confident) readout
by a wide margin (Gini $0.26$, highest cutoff $z$) and the most accurate
(P@$L/2$ $0.78$): its confidence is warranted. The single best head is the least
concentrated (Gini $0.12$) and least accurate. The perturbation readouts
(logit-/repr-/WtW-CJ) sit between in sharpness --- but on the $d'$
label-separation diagnostic they are marginally \emph{higher} than fusion
(repr-CJ $1.35$ vs.\ $1.24$): they elevate true contacts more evenly across the
map rather than concentrating signal on the top pairs. Fusion's advantage is
therefore decisiveness (concentration) plus accuracy, not raw discriminability
--- which is exactly why $d'$ is reported as a check rather than as the
confidence measure.

\begin{table}[h]
\centering
\small
\caption{\textbf{Confidence and discriminability by method} (ESM-2-650M, Zhang
eval-200, $N{=}40$ length-stratified; mean (sem)). \emph{Confidence} = sharpness
(cutoff $z$, Gini). The remaining columns are discriminability diagnostics, not
confidence: top-$L/2$ long precision, $d'$ label separation, and top-decile
enrichment. Bold marks the per-column best.}
\label{tab:confidence}
\begin{tabular}{lccccc}
\toprule
Method & cutoff $z$ & Gini & P@$L/2$ & $d'$ & top-dec. \\
\midrule
Fusion (top-10) & \textbf{6.79} (.19) & \textbf{0.259} (.009) & \textbf{0.778} (.012) & 1.240 (.018) & \textbf{0.120} (.005) \\
repr-CJ          & 6.55 (.19) & 0.159 (.007) & 0.748 (.013) & \textbf{1.352} (.029) & 0.117 (.006) \\
WtW-CJ           & 6.41 (.18) & 0.153 (.007) & 0.744 (.013) & 1.334 (.028) & 0.115 (.005) \\
logit-CJ         & 6.46 (.19) & 0.167 (.007) & 0.729 (.015) & 1.271 (.027) & 0.117 (.005) \\
Top-1 head       & 5.52 (.16) & 0.117 (.004) & 0.698 (.015) & 0.916 (.015) & 0.114 (.005) \\
\bottomrule
\end{tabular}
\end{table}

\section{Head-precision profile on ProGen2 vs ESM-2-650M}
\label{app:a2}

\begin{figure}[h]
  \centering
  \includegraphics[width=0.7\linewidth]{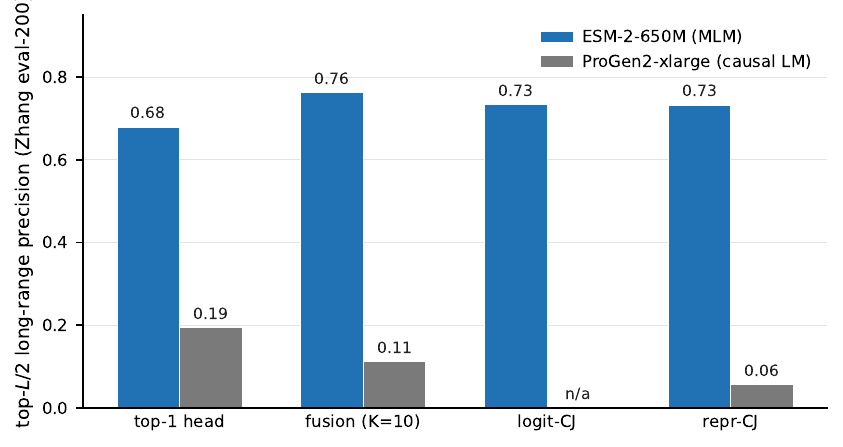}
  \caption{Contact-extraction methods on a causal LM
  (ProGen2-xlarge, next-token training) vs.\ a comparable masked
  language model (ESM-2-650M). Every method --- top-1 attention
  head, naive-mean fusion at $K{=}10$, logit-CJ \citep{zhang2024cj},
  and repr-CJ (Section~\ref{sec:repr-cj-method}) --- gives a
  substantially smaller top-$L/2$ long-range precision on ProGen2;
  logit-CJ is undefined for the causal LM and is shown as
  ``n/a''.}
  \label{fig:progen2-summary}
\end{figure}

\begin{figure}[h]
\centering
\includegraphics[width=0.95\linewidth]{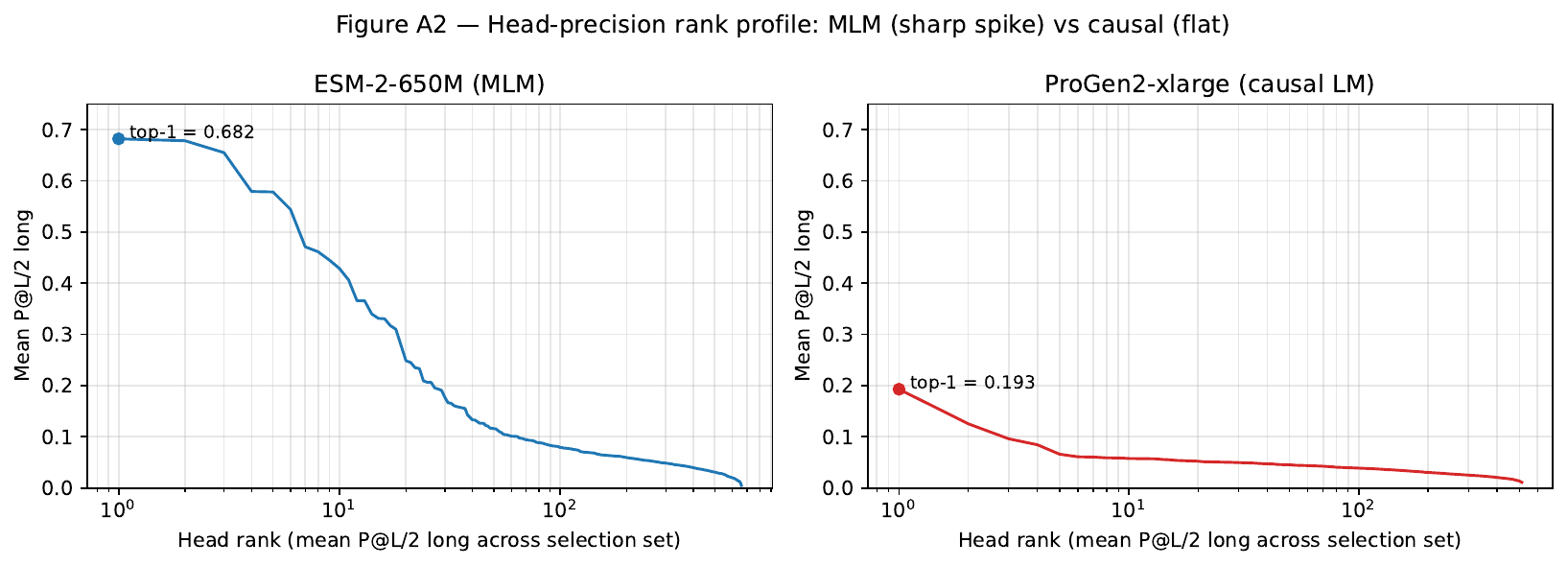}
\caption{\textbf{Head-precision rank profile: MLM (sharp spike) vs causal
(flat).} Mean P@$L/2$ long sorted by head rank, on log-rank axis
(ESM-2-650M and ProGen2-xlarge, both on Zhang eval-200). ESM-2-650M
shows the contact-head cluster spike; ProGen2-xlarge does not.}
\label{fig:a2}
\end{figure}

Figure~\ref{fig:a2} renders the mean per-(layer, head) attention-precision
profile, sorted by precision rank, for ESM-2-650M (MLM) and
ProGen2-xlarge (causal LM), both on Zhang eval-200. ESM-2-650M
shows the precision spike at low rank
that characterizes the ``contact-head cluster'' of MLM architectures
\citep{bhattacharya2021attention}: the top-ranked head reaches mean precision
$\approx 0.68$ on eval-200, with a clean drop after the top
$\sim$30 heads. ProGen2-xlarge's profile is flat: the top-ranked head
reaches only $\approx 0.19$ and the precision curve does not exhibit a
spike. This is the mechanistic signature behind the categorical scope
boundary discussed in Section~\ref{sec:causal}: causal masking removes the
gradient pressure for any attention head to specialize on bidirectional
pair structure.

\section{Depth-resolved mean-ablation probe}
\label{app:depth-probe}

To test where in the network the contact-relevant pair information
lives, we ran a sustained mean-ablation depth sweep on ESM-2-650M.
For each protein in Zhang eval-200, we replaced the residual at
every position at layer $\ell \in \{16, 24, 28, 32\}$ with the
corpus mean residual computed from a disjoint reference sample,
propagated the ablated residual through subsequent layers,
then ran Standard CJ on top of the ablated network. The
pre-specified falsification criterion: if contact signal is
localized to a single layer or layer band, we should see a
non-monotonic profile (large drop at the localized depth, small
drops elsewhere). Empirically the profile is
\textbf{strictly monotonic} across the 16-layer window we tested
(Table~\ref{tab:depth-probe}). Each ablation closer to mid-network
removes incrementally more of the contact signal.

\begin{table}[h]
\centering
\small
\caption{Depth-resolved mean-ablation probe on ESM-2-650M Zhang
eval-200 ($N{=}200$). Single-layer = ablate residual at $\ell$ only; sustained
= ablate at $\ell$ and re-apply at all subsequent layers. The two
agree within $0.4$pp on each row, so the monotonic accumulation
reflects the network's reliance on residual content at the ablation
depth, not the propagation arithmetic of sustained replacement.}
\label{tab:depth-probe}
\begin{tabular}{lcc}
\toprule
Ablation depth $\ell$ & Sustained CJ ($\text{P}@L/2$ long) & Single-layer CJ \\
\midrule
unablated   & 0.7330 & 0.7330 \\
$\ell = 32$ & 0.6801 ($-5.3$pp) & 0.6811 \\
$\ell = 28$ & 0.6704 ($-6.3$pp) & --- \\
$\ell = 24$ & 0.6591 ($-7.4$pp) & --- \\
$\ell = 16$ & 0.6046 ($-12.8$pp) & 0.6032 \\
\bottomrule
\end{tabular}
\end{table}

The interpretation: contact-relevant pair structure is built up
progressively through the residual stream rather than localized at a
single layer, with the late-layer band $\ell \in \{28, 32\}$
contributing the final $\sim$5pp of CJ's contact precision. In-manifold
mean replacement is cleaner than zero-ablation, which collapses the
network into an out-of-distribution regime and gives uninformative
drop magnitudes; we use mean replacement throughout.

\section{Per-test wall-clock and selection-set transfer}
\label{app:a5}

Table~\ref{tab:a5} reports the per-protein \emph{paired} wall-clock and peak
GPU memory behind the compute-economy figure (Figure~\ref{fig:cost},
Section~\ref{sec:cost}): CJ and fusion timed on the \emph{same} proteins on
identical A100-80GB hardware (bf16), as medians over the evaluation draw. Per
protein, CJ scales with $L$ through its $\sim 19L$ forward passes while fusion
runs a single forward (still $O(L^2)$, but $2$--$3$ orders of magnitude below
CJ), for median speedups $\sim 150$--$1600\times$ on Zhang eval-200. The
one-time 10-protein head probe is $\sim$10 eager attention forwards (reported
as a forward count rather than wall-clock: it scores all heads and was not
separately timed).

\begin{table}[h]
\centering
\small
\caption{\textbf{Per-protein \emph{paired} wall-clock and peak memory
(A100-80GB, bf16, matched batching).} CJ and fusion are timed on the
\emph{same} proteins on identical hardware; all wall-clock values are
medians over the evaluation draw. ``Speedup'' is the median per-protein
CJ/fusion wall-clock ratio. ``Peak mem'' is the maximum GPU memory (GiB)
over the draw (CJ\,/\,fusion). Fusion's single forward grows mildly with
$L$ (Figure~\ref{fig:cost}a) but stays $2$--$3$ orders of magnitude below
CJ's $\sim 19L$ forwards. The CASP14-FM row is over the $N{=}14$ targets
both methods can run; ESM-2's 1024-token context excludes the two longest
CASP14 targets ($L{=}1372, 2180$) from both methods, so we make no
$L>1024$ timing claim. Fusion's one-time 10-protein head probe is 10 eager
forwards (a forward count; not separately timed).}
\label{tab:a5}
\begin{tabular}{llrrrr}
\toprule
Variant & Dataset & CJ s/prot & Fusion s/prot & Speedup & Peak mem (CJ\,/\,fus) \\
\midrule
ESM-2-35M  & Zhang eval-200 & 4.7  & 0.032 & $150\times$  & 2.4 / 0.5 \\
ESM-2-150M & Zhang eval-200 & 12.5 & 0.046 & $280\times$  & 3.7 / 1.8 \\
ESM-2-650M & Zhang eval-200 & 30.1 & 0.050 & $604\times$  & 15.4 / 4.2 \\
ESM-2-3B   & Zhang eval-200 & 87.3 & 0.054 & $1626\times$ & 44.9 / 19.5 \\
ESM-2-3B   & CASP14-FM      & 20.1 & 0.044 & $466\times$  & 46.8 / 30.6 \\
\bottomrule
\end{tabular}
\end{table}

\paragraph{Selection-set robustness.}
The per-architecture $K$ choice is not overfit to the labeled selection set:
Appendix~\ref{app:selection-size} shows the top-10 head ranking is recovered
from as few as 10 proteins with no downstream precision penalty (the
per-architecture holdout-$K$ transfer to a disjoint draw, omitted here for
space, gave $\Delta_{\text{holdout}-\text{in-dist}} \geq 0$ on every cell that
runs; the producing script is retained in the code release).

\section{Selection-set size sweep}
\label{app:selection-size}

How small can the labeled selection set get before the head ranking
degrades? We subsample $N_{\mathrm{sub}} \in \{3, 5, 10, 15, 20, 30,
50\}$ proteins from a 50-protein labeled pool (20 random draws per
$N_{\mathrm{sub}}$; deterministic at $N_{\mathrm{sub}}{=}50$), rank
heads by mean per-protein precision on the subsample, pick the top
$K{=}10$, and evaluate naive-mean fusion on a disjoint 50-protein
holdout. Table~\ref{tab:selection-size} reports the Jaccard
overlap between the subsampled top-10 and the full-50 top-10, and the
resulting holdout fusion precision.

\begin{table}[h]
\centering
\small
\caption{\textbf{Selection-set size sweep on ESM-2-650M.} Jaccard
overlap between the top-$K{=}10$ heads picked from a subsample of
$N_{\mathrm{sub}}$ labeled proteins and from the full 50-protein set,
plus the downstream fusion precision on a disjoint 50-protein
holdout ($N{=}50$). Mean $\pm$ std over 20 random draws. \textbf{As
few as 10--15 proteins recover the full-50 head ranking and downstream
precision.}}
\label{tab:selection-size}
\begin{tabular}{rccc}
\toprule
$N_{\mathrm{sub}}$ & Jaccard vs full-50 & Fusion P@$L/2$ long & Range \\
\midrule
3  & $0.909 \pm 0.091$ & $0.786 \pm 0.005$ & $[0.782,\, 0.795]$ \\
5  & $0.955 \pm 0.079$ & $0.784 \pm 0.003$ & $[0.782,\, 0.791]$ \\
10 & $0.991 \pm 0.040$ & $0.783 \pm 0.002$ & $[0.782,\, 0.791]$ \\
15 & $1.000 \pm 0.000$ & $0.782 \pm 0.000$ & $[0.782,\, 0.782]$ \\
20 & $0.991 \pm 0.040$ & $0.783 \pm 0.002$ & $[0.782,\, 0.791]$ \\
30 & $1.000 \pm 0.000$ & $0.782 \pm 0.000$ & $[0.782,\, 0.782]$ \\
50 & $1.000$ & $0.782$ & --- \\
\bottomrule
\end{tabular}
\end{table}

The ranking is remarkably stable: at $N_{\mathrm{sub}}{=}15$ every draw
recovers the identical top-10 as the full 50-protein ranking (Jaccard
$= 1.0$), and even at $N_{\mathrm{sub}}{=}3$ the overlap averages
$0.91$ with no downstream precision penalty --- the subsampled
top-10 occasionally picks marginally different heads that score
comparably on the holdout. The contact-head cluster is sufficiently
prominent that a handful of labeled proteins identifies it. This sweep
is what licenses the 10-protein selection set used for the in-distribution
Zhang eval-200 results; the 50-protein budget elsewhere (the matched-budget
$L_1$-LR control) is conservative by comparison.

\section{Scaling beyond the context window via stitching}
\label{app:stitch}

\begin{figure}[h]
  \centering
  \includegraphics[width=\linewidth]{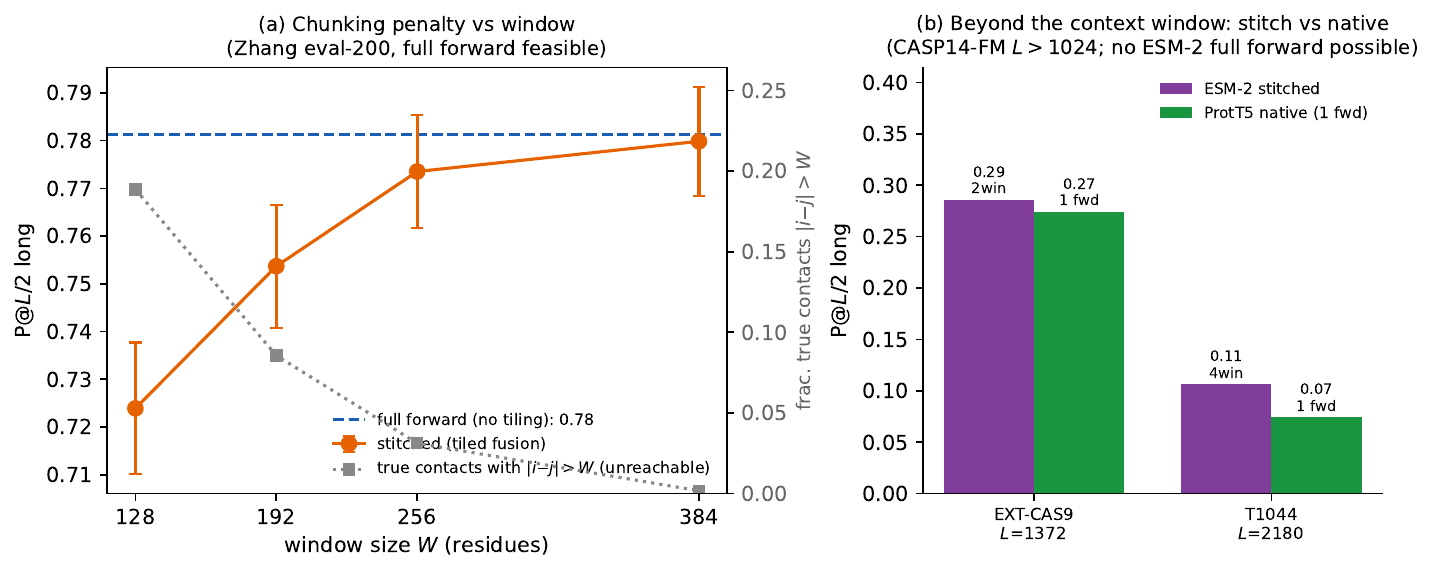}
  \caption{\textbf{Stitched fusion scales past the context window.}
  (a)~Chunking-penalty ablation on Zhang eval-200 (full forward feasible as
  baseline, blue dashed): stitched top-$L/2$ long precision (orange)
  approaches the full forward as the window $W$ grows, and the residual loss
  tracks the fraction of true contacts with $|i-j|>W$ (grey, right axis) ---
  the inherent reach limit of a fixed window, not a stitching artifact.
  (b)~The two CASP14-FM targets longer than ESM-2's $1{,}024$ context
  (EXT-CAS9 $L{=}1372$, T1044 $L{=}2180$): no ESM-2 forward can process them,
  yet \emph{stitched} ESM-2 fusion (2/4 windows, purple) and \emph{native}
  ProtT5-XL fusion (one forward, green; ProtT5's relative-position attention
  has no fixed absolute-position cap) recover comparable contacts --- two routes past the
  context window, both $\sim$1\,s where CJ would need $\sim 19L$ forwards.}
  \label{fig:stitch}
\end{figure}

The single-forward design caps fusion at the backbone's context length
(ESM-2: $1{,}024$ tokens). Longer chains can still be read by tiling the
sequence into overlapping windows ($W\leq1024$, $50\%$ overlap), running
fusion in each, and stitching the per-window maps (overlap-averaged). With
the $50\%$ stride, pairs with $|i-j|\leq W/2$ always share a window, pairs
between $W/2$ and $W$ are alignment-dependent, and only pairs with
$|i-j|>W$ are unreachable.
Two findings (Figure~\ref{fig:stitch}).
\emph{(i)~Tiling is nearly lossless at a moderate window.} On Zhang
eval-200, where the full forward is feasible as a baseline, stitching costs
$0.8$pp of top-$L/2$ long precision at $W{=}256$ (two windows) and only
$0.1$pp at $W{=}384$; the penalty grows for smaller windows ($5.7$pp at
$W{=}128$, ${\sim}4.6$ windows) and tracks the fraction of true contacts
with $|i-j|>W$ ($3.1\%$ at $W{=}256$, $18.9\%$ at $W{=}128$). Long-range
contacts cluster within moderate sequence separation, so a moderate window
captures almost all of them and stitching artifacts are second-order.
\emph{(ii)~Long proteins are reachable two ways.} On the two CASP14-FM
targets that exceed ESM-2's context (EXT-CAS9, $L{=}1372$; T1044,
$L{=}2180$) --- which no ESM-2 forward can process --- \emph{stitched} ESM-2
fusion recovers contacts at top-$L/2$ long precision $0.29$ and $0.11$ (2
and 4 windows). Alternatively, a \emph{backbone whose positional encoding extends past a
fixed window needs no stitching}: ProtT5-XL (T5 relative-position attention,
no absolute-position table) reads both targets in a \emph{single} forward
($0.9$--$1.3$\,s, $<15$\,GB) at comparable precision ($0.27$ and $0.07$;
Figure~\ref{fig:stitch}b), and ESMC's $2{,}048$-token rotary context
likewise exceeds ESM-2's $1{,}024$. These backbones are not unbounded ---
ProtT5 was pretrained around $512$ residues and ESMC at $2{,}048$, full
self-attention cost still grows with $L$, and the longest inputs run beyond
the training distribution --- but their positional encodings accept these
targets directly where a fixed absolute-position model (ESM-1b) cannot. So a protein
beyond a backbone's context can be handled either by stitching a strong
short-context model or by running a long-context one natively; on these
targets the two are comparable, with ESM-2's stronger contact signal
slightly ahead even through the stitch. Either way the cost is $\sim$1\,s
of forwards, where a tiled Categorical Jacobian would cost $\sim 19W$ per
window (tens of thousands of forwards total): the compute gap fusion opens
at single-protein scale only widens for long chains. These two targets are
a proof of concept; a quantitative $L{>}1024$ benchmark (a dedicated eval
set) is the natural next step.

\end{document}